\pdfoutput=1
\documentclass{article}

\usepackage{arxiv}

\usepackage[utf8]{inputenc} % allow utf-8 input
\usepackage[T1]{fontenc}    % use 8-bit T1 fonts
\usepackage{hyperref}       % hyperlinks
\usepackage{url}            % simple URL typesetting
\usepackage{booktabs}       % professional-quality tables
\usepackage{amsmath}            % blackboard math symbols
\usepackage{amssymb}        % additional AMS symbols
\usepackage{amsfonts}       % blackboard math symbols
\usepackage{nicefrac}       % compact symbols for 1/2, etc.
\usepackage{microtype}      % microtypography
\usepackage{lipsum}		% Can be removed after putting your text content
\usepackage{graphicx}
\usepackage[numbers]{natbib}
\usepackage{doi}

\usepackage{hyperref}
\usepackage{url}

\usepackage{graphicx}

\usepackage{booktabs}

\usepackage{subcaption}

\usepackage{multirow}

\usepackage[T1]{fontenc}
\usepackage{tikz}
\usetikzlibrary{positioning,calc}
\usepackage{listings}
\usepackage{xcolor}

\lstdefinestyle{pythonbox}{
    language=Python,
    basicstyle=\ttfamily\scriptsize,
    keywordstyle=\bfseries,
    commentstyle=\itshape,
    stringstyle=\itshape,
    showstringspaces=false,
    columns=fullflexible,
    keepspaces=true,
    breaklines=true,
    breakatwhitespace=false,
    tabsize=4,
    frame=none
}

\usepackage{enumitem}

\title{EvoMO-SR: Multiobjective LLM-based Evolution of Symbolic Expressions with substructure guidance}

\date{}

\author{ {\hspace{1mm}Cristina~Rossetti}\thanks{Corresponding author.} \\
	Department of Automation and Computer Science\\
	Polytechnic Institute of Turin\\
	Corso Duca degli Abruzzi, 24, Torino, 10129, Italy \\
	\texttt{cristina.rossetti@polito.it} \\
    	\And
	{\hspace{1mm}Anna V.~Kononova} \\
	LIACS\\
    Leiden University\\
	Einsteinweg 55, 2333 CC, Leiden, The Netherlands\\
	\texttt{a.kononova@liacs.leidenuniv.nl} \\
    	\And
	{\hspace{1mm}Thomas ~Bäck} \\
	LIACS\\
    Leiden University\\
	Einsteinweg 55, 2333 CC, Leiden, The Netherlands\\
	\texttt{T.H.W.Baeck@liacs.leidenuniv.nl} \\
    	\And
	{\hspace{1mm}Fei ~Liu} \\
	University of Zurich \& ETH Zurich\\
	\texttt{fei.liu@my.cityu.edu.hk} \\
	\And
	{\hspace{1mm}Niki~Van Stein} \\
	LIACS\\
    Leiden University\\
	Einsteinweg 55, 2333 CC, Leiden, The Netherlands\\
	\texttt{n.van.stein@liacs.leidenuniv.nl} \\
}

\renewcommand{\shorttitle}{EvoMO-SR}

\hypersetup{
pdftitle={A template for the arxiv style},
pdfsubject={q-bio.NC, q-bio.QM},
pdfauthor={David S.~Hippocampus, Elias D.~Striatum},
pdfkeywords={First keyword, Second keyword, More},
}

\begin{document}
\maketitle

\begin{abstract}
	Symbolic Regression (SR) is a data-driven method for scientific discovery which searches for interpretable analytical relationships within data. Recently, Large Language Models (LLMs) have also had a significant impact on scientific discovery, enabling the automation of various stages of the process. For these reasons, the possibility of harnessing the embedded scientific knowledge and programming capabilities of LLMs to solve SR tasks has emerged, showing promising performance compared with traditional methods. We propose EvoMO-SR, a novel LLM-driven SR framework in which the LLM generates equation skeletons, with their coefficients fitted separately by an external optimizer. The framework includes a multi-objective survival selection which controls bloating by balancing accuracy and complexity, and a substructure guidance mechanism which mutates expressions with candidate reusable building blocks. EvoMO-SR achieves the best accuracy in seven of the eight in-domain and out-of-domain settings for LSR-Synth, using a small LLM model, i.e., Llama-3.1-8B-Instruct. We also evaluated structural recovery through two symbolic accuracy metrics based on canonicalized subtree overlap and term matching, showing that our method has a greater probability of recovering highly accurate symbolic structures.
\end{abstract}

% keywords can be removed

\section{Introduction}
Large Language Models (LLMs) are increasingly being used for scientific discovery tasks, given their vast embedded knowledge and ability to understand and generate natural language \citep{reddy2025towards, chen2025ai4research, song2025evaluating, zheng2025automation}. 
In this context, LLMs can be used at various stages, from literature analysis to the generation of new ideas, right through to more advanced applications such as the use of agents to automate the scientific discovery process, thereby minimizing the need for human intervention \citep{zheng2025automation}.
One way in which scientific discovery is enhanced is through data-driven discovery, thanks to the growing availability of extensive amounts of data and tools capable of processing it and extracting useful insights \citep{reddy2025towards}. 
Symbolic Regression (SR), also known as \textit{equation discovery}, can be framed as a data-driven discovery method for scientific research \citep{shojaee2025llm} as it enables the automatic derivation of analytical relationships, revealing complex and hidden patterns within the data. Given that the resulting models make explicit the mathematical relationship between inputs and outputs, SR has gained increasing prominence in scientific discovery due to its greater interpretability compared to black-box models
\citep{makke2024interpretable}.
Since its advent, SR has been based primarily on Genetic Programming (GP) methods \citep{koza1994genetic}, e.g., pySR \citep{cranmer2023interpretable}, GPGomea \citep{virgolin2021improving}, which generate formulas in the form of expression trees, alternating between crossover and mutation operators within evolutionary loops. Moreover, methods based on deep learning have also been proposed, employing for instance deep reinforcement learning \citep{petersen2019deep} and transformers \citep{kamienny2023deep}.
Recently, the widespread adoption of LLMs has facilitated their integration with SR, given their extensive embedded scientific knowledge, and the ability to easily incorporate domain-specific priors \citep{dong2025recent,shojaee2025llm}. LLMs can act as hypothesis generators, refining expressions in evolutionary-like loops, and can be prompted with contextual information, user hints, abstract concepts, and formulas or ideas extracted in past generations \citep{shojaee2025llmsr, guo2026coevo, srwithlearnedconceptlibrary}.  

In this paper, we propose EvoMO-SR, an LLM-based SR framework that generates expressions in an evolutionary loop (overview of the architecture shown in Fig. \ref{architecture}). The LLM is responsible for generating expression skeletons in the form of Python code, similar to \citep{shojaee2025llmsr}, while the estimation of the coefficients is performed by a specific optimizer, in our case L-BFGS-B \citep{zhu1997algorithm}.
Generation occurs within an explicit evolutionary loop in which the LLM is provided with a context prompt and specific instructions to perform the task. 
Since one of the major issues already well recognized in the field of SR involves the risk of bloating, i.e., the phenomenon in which formulas uncontrollably grow in their structure \citep{kronberger2024symbolic}, the framework implements a multi-objective survival selection \citep{deb2011multi}, balancing accuracy and complexity. Survivors are selected from the Pareto frontier generated using the Non-Dominated Sorting and Crowding Distance algorithm \citep{deb2002fast, smits2005pareto}. Additionally, to help identifying candidate reusable building blocks, EvoMO-SR manages an archive of functional substructures that are extracted during evolution from generated offspring individuals. A substructure can be randomly selected from the archive based on a pre-defined probability in order to guide the LLM in the mutation of individuals.

We evaluated EvoMO-SR on the LSR-Synth dataset, which includes 129 problems, from LLM-SRBench \citep{shojaee2025llm} and on custom benchmark problems \citep{shojaee2025llmsr}. By incorporating synthetic terms into known scientific models, both datasets were designed to mitigate the issue of memorization by the LLM, thereby encouraging the LLM to effectively perform data-driven SR. To this purpose, we also evaluated our method on four datasets created from randomly generated equations. Main results, obtained by using Llama-3.1-8B-Instruct, show that EvoMO-SR outperforms SR baselines in most cases, both in in-domain (ID) and in out-of-domain (OOD) settings. In details, EvoMO-SR achieves the lowest aggregate Normalized Mean Squared Error (NMSE) among the evaluated methods in seven of the eight LSR-Synth domain-split comparisons, including all four OOD. This is notable as it demonstrates that our method performs well even with a relatively small model. In addition, we also assessed the symbolic accuracy of the resulting expressions, i.e., their structural similarity to the corresponding ground truth, through two metrics. Although none of the methods achieves satisfactory symbolic accuracy, especially for LSR-Synth, EvoMO-SR showed a greater probability of recovering highly accurate symbolic structures.

The main contributions are as follows:

\begin{itemize}
    \item We introduce EvoMO-SR, a framework for SR that employs LLMs in an evolutionary strategy to generate equation skeletons, while coefficients are fitted separately using an external optimizer. The survival mechanism is based on a multi-objective strategy to balance numerical accuracy and complexity, while reusable functional substructures are stored in a dynamic archive to enrich subsequent prompts.

    \item We evaluate EvoMO-SR on the full LSR-Synth benchmark and additional oscillator problems tailored to mitigate the LLM memorization issue. We compared results against LLM-SR \citep{shojaee2025llmsr}, LaSR \citep{srwithlearnedconceptlibrary} and state-of-the-art SR baselines. The framework achieves the lowest aggregate NMSE in seven of the eight ID and OOD comparisons over LSR-Synth. Furthermore, we compare our method with LLM-SR on the four datasets created from randomly generated functions, consistently showing better NMSE for ID and OOD and better symbolic accuracy in most cases.

    \item We enrich the evaluation with a symbolic accuracy analysis based on two deterministic metrics. Although the results show that no method is consistently able to achieve high symbolic accuracy scores, EvoMO-SR demonstrates a slightly greater ability to retrieve expressions that are very close to the ground truth.

\end{itemize}

\section{Related Works}
\label{sec:related_work}
SR is a supervised machine learning method that aims to retrieve the mathematical expression that best fits a given dataset. The growing popularity of SR is due to its ability to uncover complex hidden patterns without making any prior assumptions about the functional form on the data, and to reveal the analytical relationships between features, thereby enhancing interpretability \citep{dong2025recent, makke2024interpretable}. While SR methods are mainly based on GP \citep{cranmer2023interpretable,virgolin2021improving} and deep learning \citep{kamienny2023deep, petersen2019deep}, the advent of LLMs has had an impact on the development of innovative methods for SR. One of the most representative methods is LLM-SR \citep{shojaee2025llmsr}, which employs LLMs to generate expression skeletons in an evolutionary-like loop, with the fitting of coefficients left to off-the-shelf optimizers. The LLM is prompted with high-performing candidates sampled via a multi-island approach from a dynamic experience buffer, prior knowledge of the problem, and task specification. The experience buffer stores diverse promising candidates, while the sampling strategy involves the selection of shorter programs in order to mitigate bloating issues. 
In contrast to EvoMO-SR, LLM-SR does not implement an explicit population-based evolutionary algorithm with parent-offspring survival selection. Instead, its evolutionary component mainly operates through the management and fitness-biased sampling of complete candidate expressions from the experience buffer.
\cite{wang2025drsr} extend LLM-SR with data-aware insights and inductive idea extraction, which summarizes positive, negative, and invalid generation outcomes into reusable heuristics. These ideas are stored in a dynamic library and used to enrich subsequent prompts. \cite{zhang2026llm} explore the use of LLMs in SR for the automatic design of selection operators to enforce semantic guidance and mitigate bloating. Their framework evolves selection strategies through LLMs using in-context learning to improve evolutionary SR algorithms. Each strategy is evaluated through inner SR runs across multiple datasets and selection operators survival is based on a multi-objective mechanism which balances downstream fitness and operator code length.
Another relevant work is LaSR \citep{srwithlearnedconceptlibrary}, a framework which enhances GP search of equations with natural language concepts generated via LLMs. Building on PySR, LaSR does not use LLMs as its primary equation generator, but it alternates initialization, mutation, and crossover operators with their LLM variants, extracting and evolving abstract concepts from high-performing candidates and reusing them in the equation search. LaSR explicitly adds a complexity penalty term in the fitness function, and selects Pareto-optimal solutions based on dataset loss and syntactic complexity to extract abstract ideas. However, the evolutionary process is mainly guided by traditional GP and the archive used for guidance stores natural language ideas instead of explicit mathematical pieces. 

% CoEvo \citep{guo2026coevo} employs a dynamic library of ideas that are extracted, refined and reused by the LLM. Unlike LaSR, CoEvo directly uses an LLM within an evolutionary loop, refining multi-format symbolic formulas with mutation and crossover operators and reusing the stored ideas. CoEvo introduces a negative crossover operator to encourage the LLM to generate solutions different from the selected ideas, thus enhancing diversity. However, its selection mechanism remains limited to the candidates' score.  
% Our work integrates LLMs into an evolutionary loop that introduces explicit mechanisms both for diversity preservation and multi-objective optimization.

\section{Methodology}
\label{sec:methodology}
\subsection{Problem Formulation}
The task of Symbolic Regression (SR), also known as \textit{equation discovery}, is to find the function that best fits a given dataset.
More precisely, its goal is, given a dataset \(\mathcal{D}= \{(x_i,y_i)\}_{i=1}^n\) where \(x_i \in \mathbb{R}^d\) is the feature vector and \(y_i \in \mathbb{R}\) is the related target, to learn a symbolic functional structure \(\phi(\cdot;\hat{\Theta})\) and corresponding parameter values \(\hat{\Theta}\) such that \(\hat{f}(x) = \phi(x;\hat{\Theta}):\mathbb{R}^d \rightarrow \mathbb{R}\) and \(\hat{f}(x_i) \approx y_i\), for \(i=1,\dots,n\).
Since SR methods typically generate multiple analytical models for the same problem, they can be seen as \textit{hypotheses generating machines} \citep{kronberger2024symbolic}. In our framework, we generate hypotheses using an LLM in the form of equation skeletons, leaving the estimation of parameters to specific optimizers. Formally, the LLM generates a set of candidate equation skeletons \(\Phi = \{\phi_{j}\}_{j=1}^M\), where each \(\phi_{j}\) represents a symbolic expression with unestimated numerical parameters. The parameters of the equations are estimated separately by optimizing a fitting loss \(\mathcal{L}\) over \(\mathcal{D}\). The set of fitted candidate hypotheses is then \(\{\hat{f}_j(x)\; | \; \hat{f}_j(x) = \phi_j(x;\hat{\Theta}_j)\}_{j=1}^M\).

\subsection{EvoMO-SR methodology}
\begin{figure}[t]
\centering
\includegraphics[width=1.0\linewidth]{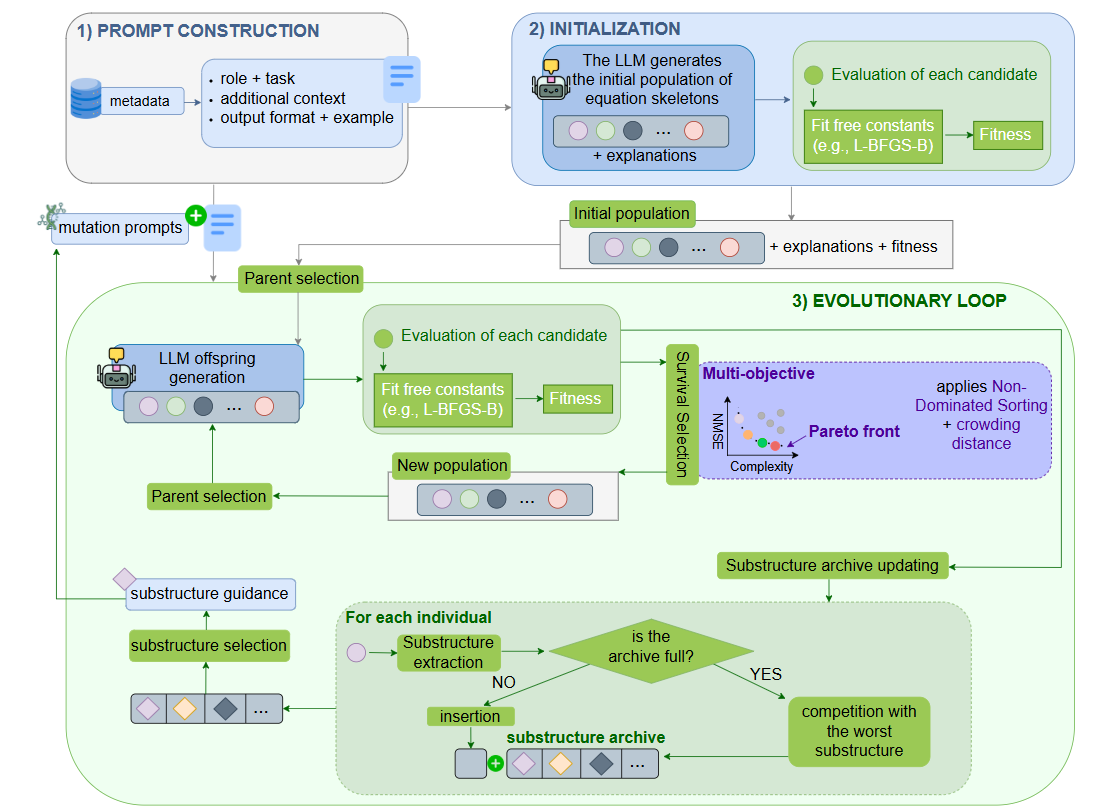}
  \caption{EvoMO-SR main components. \textbf{1) Prompt Construction}: the first step includes the construction of the prompt by role specification, task description, additional context based on considered dataset, and output and example format. \textbf{2) Initialization}: the initial population is created by prompting the LLM with the constructed prompt, and each candidate is fitted and evaluated. \textbf{3) Evolutionary Loop}: in subsequent generations, the LLM is prompted with the initial prompt enriched with the description, fitness and equation of the selected candidate, plus a mutation prompt randomly selected from the pre-fixed set. To form the new population, the fitted and evaluated set of candidates faces survival selection, which is on multi-objective optimization. At the end of each generation, each individual is scanned to extract added substructures and the substructure archive is updated accordingly.}
  \label{architecture}
\end{figure}
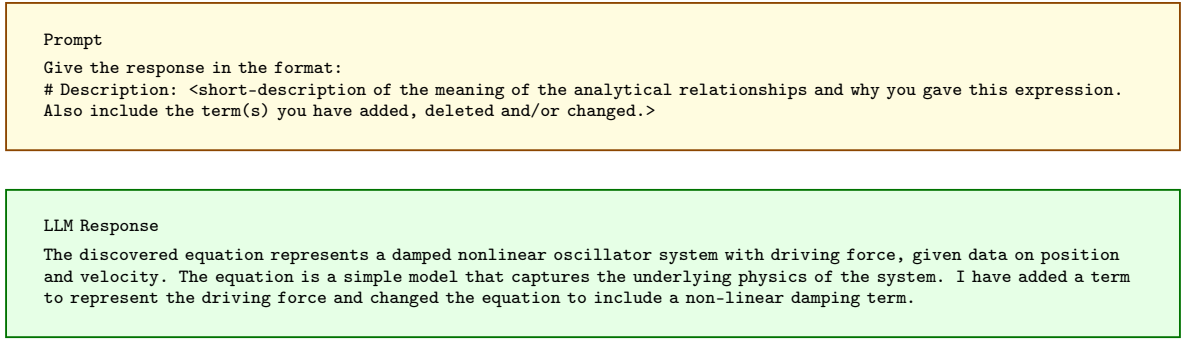
\begin{figure}[t]
\centering

\begin{tikzpicture}[
    box/.style={
        text width=0.88\linewidth,
        align=left,
        inner xsep=5mm,
        inner ysep=4mm,
        line width=0.7pt,
        font=\ttfamily\scriptsize
    },
    prompt/.style={
        box,
        fill=yellow!15,
        draw=orange!55!black
    },
    response/.style={
        box,
        fill=green!10,
        draw=green!45!black
    }
]

\node[prompt] (p) {
\textbf{Prompt}\\[1mm]
Give the response in the format:

\# Description: <short-description of the meaning of the analytical
relationships and why you gave this expression. Also include the term(s)
you have added, deleted and/or changed.>
};

\node[
    response,
    below=5mm of p
] (r) {
\textbf{LLM Response}\\[1mm]
The discovered equation represents a damped nonlinear oscillator system
with driving force, given data on position and velocity. The equation is a
simple model that captures the underlying physics of the system. I have
added a term to represent the driving force and changed the equation to
include a non-linear damping term.
};

\end{tikzpicture}

\caption{Example of the feedback generated by the LLM.}
\label{fig:prompt-descr}
\end{figure}

EvoMO-SR performs SR employing an LLM to generate and refine equation skeletons within an evolutionary loop (see main components of the architecture in Fig. \ref{architecture}). 
The initial population of equation skeletons \(\Phi = \{\phi_{j}\}_{j=1}^M\), where \(M\) denotes the number of solutions in the parent population, is generated based on a set of prompts \(\mathcal{P_{\text{init}}} = \{\text{role}+ \text{task}, \text{context}, \text{format}\}\) that are concatenated. The context is specific to the dataset \(\mathcal{D}\) under consideration, and is intended to support equation generation by incorporating information such as the physical meaning of the variables. The LLM’s output includes not only the equation skeleton (in the form of Python code) but also an explanation of why that specific structure was proposed. These explanations are saved as textual descriptions of each equation, alongside the proposed formula, so as to enrich subsequent prompts. Figure \ref{fig:prompt-descr} shows an example of an explanation given by the LLM for the generation of an equation skeleton. These descriptions are not exploited only to provide the LLM with useful information about previously generated candidates, but also to make the process more interpretable. In fact, in this way, we can trace the decisions made by the LLM for each generated individual, thus understanding the motivations behind the proposal or refinement of expressions.

Once the generation ends, each equation in the initial population is fitted with an external optimizer, in our case L-BFGS-B, and evaluated with a fitness metric on training data. 
At this point, the actual evolutionary loop begins, starting with the mutation of individuals selected from the initial population \(\Phi = \{\phi_{j}\}_{j=1}^M\). The parent selection strategy, which decides the individuals to mutate from the current population, can be customized with the desired mechanism. In our case, individuals are selected by a ``random with replacement'' strategy. Details on additional evolutionary components and parameter settings are available in Appendix \ref{app:add_ev_components}. An individual is mutated by explicitly prompting the LLM with a specific mutation prompt, chosen at random at each iteration (where each iteration corresponds to the mutation of one individual in a generation). In our main setting, the possibilities are whether to refine the current expression or to propose a new one. So, in order to generate the new offspring population, the LLM is prompted with a set of prompts \(\mathcal{P}_{\text{loop}}=\mathcal{P_{\text{init}}}\cup\{\mathcal{I}_j,\texttt{mutation}\}\), where \(\mathcal{I}_j\) is the current selected individual from the parent population that includes its textual description and fitness. The generated skeletons \(\Phi = \{\phi_{k}\}_{k=1}^N\), where \(N\) denotes the number of solutions in the offspring population, are then fitted and evaluated. The new population is created based on the multi-objective survival strategy.
 % When the \textit{elitism} parameter is set to \textit{True}, both parents \(\{\hat{f}_j(x)\}_{j=1}^M\) and offspring individuals \(\{\hat{f}_k(x)\}_{k=1}^N\) are included in the candidate pool for survival selection. 
On this set of individuals, the non-dominated sorting algorithm is applied to form Pareto fronts based on two opposing objectives: accuracy (i.e., NMSE) and complexity (i.e., count of nodes in the expression tree, following \cite{landajuela2022unified, la2021contemporary, cranmer2023interpretable}). 
In detail, NMSE is defined as \(\text{NMSE} =\frac{
\sum_{i=1}^{N}\left(y_i-\hat{y}_i\right)^2
}{
\sum_{i=1}^{N}\left(y_i-\bar{y}\right)^2
}\), where \(\bar y\) denotes the mean target value. Lower values indicate better predictive accuracy, with zero corresponding to a perfect fit. The complexity is computed as the number of nodes of the Python Abstract Syntax Tree (AST) of the discovered expression after coefficient optimization.
The next population is formed by taking individuals front by front until the maximum number of individuals to be retained is reached. If the number of individuals on a front exceeds the remaining space in the population being constructed, the crowding distance criterion is applied, measuring how isolated a solution is in space. The algorithm favours solutions with a higher crowding distance, in order to promote greater diversity. A Pareto Archive is maintained to keep trace of all the best trade-offs candidates found during the whole search. The new population is then given as input to the next generation. The evolutionary loop ends when the pre-fixed number of generations is reached.

At the end of each generation, each individual in the offspring population is scanned to extract added substructures. Here we refer as a \textit{substructure} an additive subtree of the input expression tree. This extraction enables to detect new functional structures added by the LLM during the evolution, thus identifying possible useful and reusable components. In order to keep trace of these substructures, the system maintains a dynamic fixed-length archive that is updated at each generation. When a new substructure is detected in an individual, it is checked whether there is space left in the archive. If yes, the new substructure \(s'\) is added to the archive; otherwise, \(s'\) competes for the insertion against the worst substructure \(s_{worst}\) in the archive. Since for each substructure is stored the NMSE of the related offspring, \(s'\) replaces  \(s_{worst}\) if and only if the NMSE related to \(s'\) is lower than the one of \(s_{worst}\).
The archive of substructures occasionally serves to enrich mutation prompts based on a probability \(p_{sub}\). When this happens, a substructure is randomly selected from the archive and is used for substructure guidance by telling the LLM to use that substructure in the mutation.

% \begin{figure}[]
% \centering
%   \centering \includegraphics[width=0.8\linewidth]{images/prompt_descr.png}

%   \caption{Prompt of the format for the description and the response of the LLM for one iteration of a run over the \texttt{oscillator1} dataset. }
%   \label{fig:prompt-descr}
% \end{figure}

\subsection{Symbolic Accuracy metrics}
\label{sec:SA_metrics}
The evaluation of the accuracy for SR problems should consider the Symbolic Accuracy (SA), i.e., how similar is the expression found to the ground truth equation.
We evaluate SA by computing the structural similarity between each discovered equation \(\hat{f}^{\star}\) and the ground-truth formula \(f^{\star}\). To this purpose, we consider two complementary structural metrics: an AST-based subtree similarity (\(SA_{AST}\)) and an additive term-based similarity (\(SA_{Term}\)).
The first metric measures the degree of structural overlap between \(\hat{f}^{\star}\) and \(f^{\star}\) with the Jaccard similarity of their subtree signatures, similarly to \citep{kahlmeyer2025scaling}. First, both \(\hat{f}^{\star}\) and \(f^{\star}\) are preprocessed to obtain comparable forms by unifying the syntax of operators, and, optionally, applying SymPy-based canonicalization\footnote{https://docs.sympy.org/latest/index.html}. Each expression is then parsed into an Abstract Syntax Tree (AST). Further normalization steps include transforming  subtraction and division operators into canonical additive/multiplicative forms, sorting of commutative operands, and replacement of numeric constants by placeholders, so that structurally identical/similar formulas with different fitted constants can be fairly compared. Finally, the set of serialized subtree signatures is extracted from the two canonical ASTs, and the \(SA_{AST}\) is computed with their Jaccard similarity. So, given \(\mathcal{T}(f)\) being the set of subtree signatures extracted from \(f\), we define the (\(SA_{AST}\)) between \(\hat{f}^{\star}\) and \(f^{\star}\) as follows:
\[SA_{AST}(\hat{f}^*,f^*)=\frac{\left| \mathcal{T}(\hat{f}^{\star}) \cap \mathcal{T}(f^{\star}) \right|}{\left| \mathcal{T}(\hat{f}^{\star}) \cup \mathcal{T}(f^{\star}) \right|}.\]

This metric ranges from $0$ to $1$, with larger values indicating greater overlap between the hierarchical structures of the two expressions. A value of 0 indicates no overlap between subtrees and a value of 1 indicates complete overlap and therefore perfect structural equivalence between the two expressions.

We also computed \(SA_{Term}\), a Jaccard similarity metric based on the matching additive building blocks (or terms) of the expressions. After the canonicalization of both $\hat{f}^{\star}$ and $f^{\star}$, the additive terms are exposed and, for each of them, variable-independent multiplicative factors are removed, so that fitted numerical coefficients do not affect the comparison.
Let $\mathcal{T}_{Term}(f)$ denote the set of normalized additive-term signatures extracted from $f$. The term-based \(SA_{Term}\) is defined as follows:

\[SA_{Term}(\hat{f}^*,f^*)=
\frac{
\left|
\mathcal{T}_{Term}(\hat{f}^{\star})
\cap
\mathcal{T}_{Term}(f^{\star})
\right|
}{
\left|
\mathcal{T}_{Term}(\hat{f}^{\star})
\cup
\mathcal{T}_{Term}(f^{\star})
\right|
}.
\]

The term-based metric therefore measures how many of the normalized additive components of the ground-truth expression are also present in the discovered equation. It ranges from $0$ to $1$, where $0$ denotes no shared normalized terms and $1$ indicates that the two expressions contain the same set of normalized additive terms. The \(SA_{Term}\) metric is less sensitive to additional or structurally different components, as it focuses on matching normalized additive terms. Consequently, it gives relatively more credit to equations that recover correct ground-truth substructures even when they also contain spurious terms, while \(SA_{AST}\) tends to penalize more severely. More details in Appendix \ref{app:symbolic_accuracy}.

\section{Experimental results}
\label{sec:exp_results}
The main experiments were conducted using Llama-3.1-8B-Instruct as the LLM backbone, with comparisons against two LLM-based baselines, namely LLM-SR \citep{shojaee2025llmsr} and LaSR \citep{srwithlearnedconceptlibrary}, and other state-of-the-art SR baselines. For each dataset, we ran EvoMO-SR for \(100\) generations, resulting in a total of 1'000 generated expressions, optimizing parameters via L-BFGS-B, and setting the model's temperature to 0.7. The dimension of the archive of substructures was set to \(10\) with \(p_{sub}=0.5\).
%Additional details and further experiments can be found in the Appendix \ref{app:impl} and \ref{sec:app-results}.

\subsection{Comparisons over LSR-Synth dataset}
\label{sec:lsr-synth-results}

\begin{table}[b]
\caption{Comparison of Median NMSE ID and OOD values between EvoMO-SR against LLM-SR and other SR baselines over LSR-Synth datasets (44 for Physics, 36 for Chemistry, 24 for Biology, 25 for Material Science) averaged across 5 runs. EvoMO-SR, LLM-SR and LaSR have been run with Llama-3.1-8B-Instruct backbone. Lower NMSE ID/OOD (\(\downarrow\)) values are better.}

\label{tab:llama-synth}
\renewcommand{\arraystretch}{1.3}
\resizebox{\columnwidth}{!}{%
\begin{tabular}{c cc cc cc cc }

\multirow{2}{*}{\textbf{Method}}                                                & \multicolumn{2}{c}{\textbf{Physics}}                    & \multicolumn{2}{c}{\textbf{Chemistry}}                  & \multicolumn{2}{c}{\textbf{Biology}}                    & \multicolumn{2}{c}{\textbf{Material Science}}           \\ \cline{2-9} 
& \multicolumn{1}{c}{\textbf{ID}  \(\downarrow\)}      & \textbf{OOD} \(\downarrow\)     & \multicolumn{1}{c}{\textbf{ID} \(\downarrow\)}      & \textbf{OOD} \(\downarrow\)    & \multicolumn{1}{c}{\textbf{ID} \(\downarrow\)}      & \textbf{OOD} \(\downarrow\)   & \multicolumn{1}{c}{\textbf{ID} \(\downarrow\)}      & \textbf{OOD} \(\downarrow\)     \\ \hline
uDSR                                                                            & \multicolumn{1}{c}{1.15e-1}          & 9.30e-2          & \multicolumn{1}{c}{1.07e-1}          & 86.11            & \multicolumn{1}{c}{1.99e-1}          & 1.29             & \multicolumn{1}{c}{6.23e-2}          & 3.28             \\

pySR                                                                            & \multicolumn{1}{c}{3.23e-1}          & 2.30e-1          & \multicolumn{1}{c}{2.63e-1}          & 180.94           & \multicolumn{1}{c}{8.47e-1}          & 366.68           & \multicolumn{1}{c}{2.50e-1}          & 25.93            \\ 
GPGomea                                                                         & \multicolumn{1}{c}{4.11e-2}          & 4.05e-2          & \multicolumn{1}{c}{1.97e-3}          & 2.59             & \multicolumn{1}{c}{2.29e-1}          & 23.52            & \multicolumn{1}{c}{1.22e-3}          & 1.22e-1          \\ 
Operon                                                                          & \multicolumn{1}{c}{1.58e-1}          & 1.51e-1          & \multicolumn{1}{c}{1.04e-2}          & 7.54             & \multicolumn{1}{c}{3.42e-1}          & 49.43            & \multicolumn{1}{c}{1.82e-2}          & 1.11             \\ 
LLM-SR                                                                          & \multicolumn{1}{c}{\textbf{4.10e-5}} & 7.65e-4          & \multicolumn{1}{c}{1.14e-5}          & 3.34e-1          & \multicolumn{1}{c}{7.23e-5}          & 5.29e-1          & \multicolumn{1}{c}{2.20e-5}          & 6.67e-3          \\ 
LaSR                                                                          & \multicolumn{1}{c}{9.69e-3} &    3.88e-2     & \multicolumn{1}{c}{3.03e-4}          &     3.41     & \multicolumn{1}{c}{2.70e-3}          &  1.21      & \multicolumn{1}{c}{2.84e-4}          &   3.47e-2      \\ \hline 
\begin{tabular}[c]{@{}c@{}}\textbf{EvoMO-SR}\end{tabular} & \multicolumn{1}{c}{5.88e-5}          & \textbf{ 5.47e-5}    & \multicolumn{1}{c}{\textbf{2.32e-6}}    & { \textbf{6.32e-3}} & \multicolumn{1}{c}{{ \textbf{9.65e-6}}} & {\textbf{1.43e-2}} & \multicolumn{1}{c}{{\textbf{1.40e-8}}}    & {\textbf{2.73e-6}}    \\ 
\end{tabular}%
}
\end{table}

Since LLM-based SR methods may rely on the memorization of well-known formulas instead of performing real data-driven SR, \cite{shojaee2025llm} proposed LSR-Synth, a dataset including a total number of 129 problems across chemistry, biology, physics, and material science domains. The ground-truth equations have been generated from the original formulas by adding new synthetic terms. The OOD split is generated taking the final 10\% of samples (500 samples) ordered by time (or temperature for material science problems), while the remaining 90\% (4500 samples) is divided into training and ID sets. 
To ensure robustness, we ran our method and each baseline 5 times and averaged the obtained results. Table \ref{tab:llama-synth} shows the averages of the median NMSE ID and NMSE OOD values for 5 runs for EvoMO-SR, LLM-SR, LaSR, and the other SR baselines. We chose well-known SR methods that have been extensively benchmarked in the SR community \citep{de2024srbench++,imai2025call}, such as pySR \citep{cranmer2023interpretable}, GPGomea \citep{virgolin2021improving}, Operon \citep{burlacu2020operon}, and uDSR \citep{landajuela2022unified}. EvoMO-SR outperforms all baselines in seven of the eight domain-split comparisons, with LLM-SR resulting the best competitor, performing slightly better in the ID split for the Physics domain.
Compared to LLM-SR, EvoMO-SR reduces aggregate OOD NMSE by factors of approximately 14 in Physics, 53 in Chemistry, 37 in Biology, and \(2.4 \times 10^3\) in Material Science. These improvements are substantial, but they do not compare results over individual problems. 
To better analyze the differences in performance between EvoMO-SR and LLM-SR, which is our main competitor, we performed paired Wilcoxon signed-rank tests using the benchmark problems as paired experimental units, considering ID and OOD performance separately for each domain. The p-values (adjusted with Holm's correction) are always below \(0.05\), except for the Physics ID dataset, indicating statistically significant differences in performance in all other cases. Detailed results are shown in Appendix \ref{app:stat_tests}.

\begin{figure}[t]
    \centering
    \includegraphics[width=0.7\linewidth]{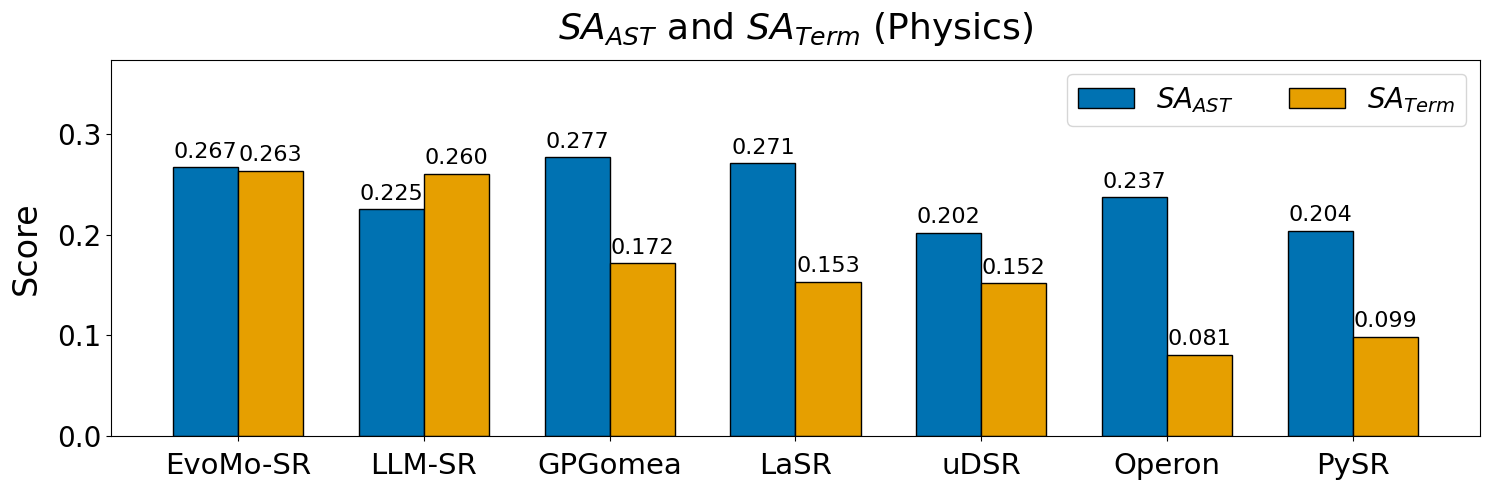}
    \caption{Comparison of \(SA_{AST}\) and \(SA_{Term}\) between EvoMO-SR and the other SR baselines over LSR-Synth Physics dataset.}
    \label{fig:SA_lsr_phys}
\end{figure}

\begin{figure}[]
    \centering
    \begin{subfigure}[]{0.55\linewidth}
        \centering
        \includegraphics[width=\textwidth,trim=10mm 120mm 5mm 0mm,clip]{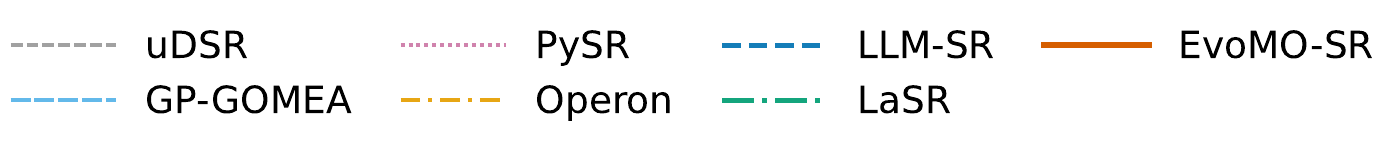}
    \end{subfigure}
    \begin{subfigure}[]{0.495\linewidth}
        \centering
        \includegraphics[width=\textwidth,trim=1mm 4mm 14mm 26mm,clip]{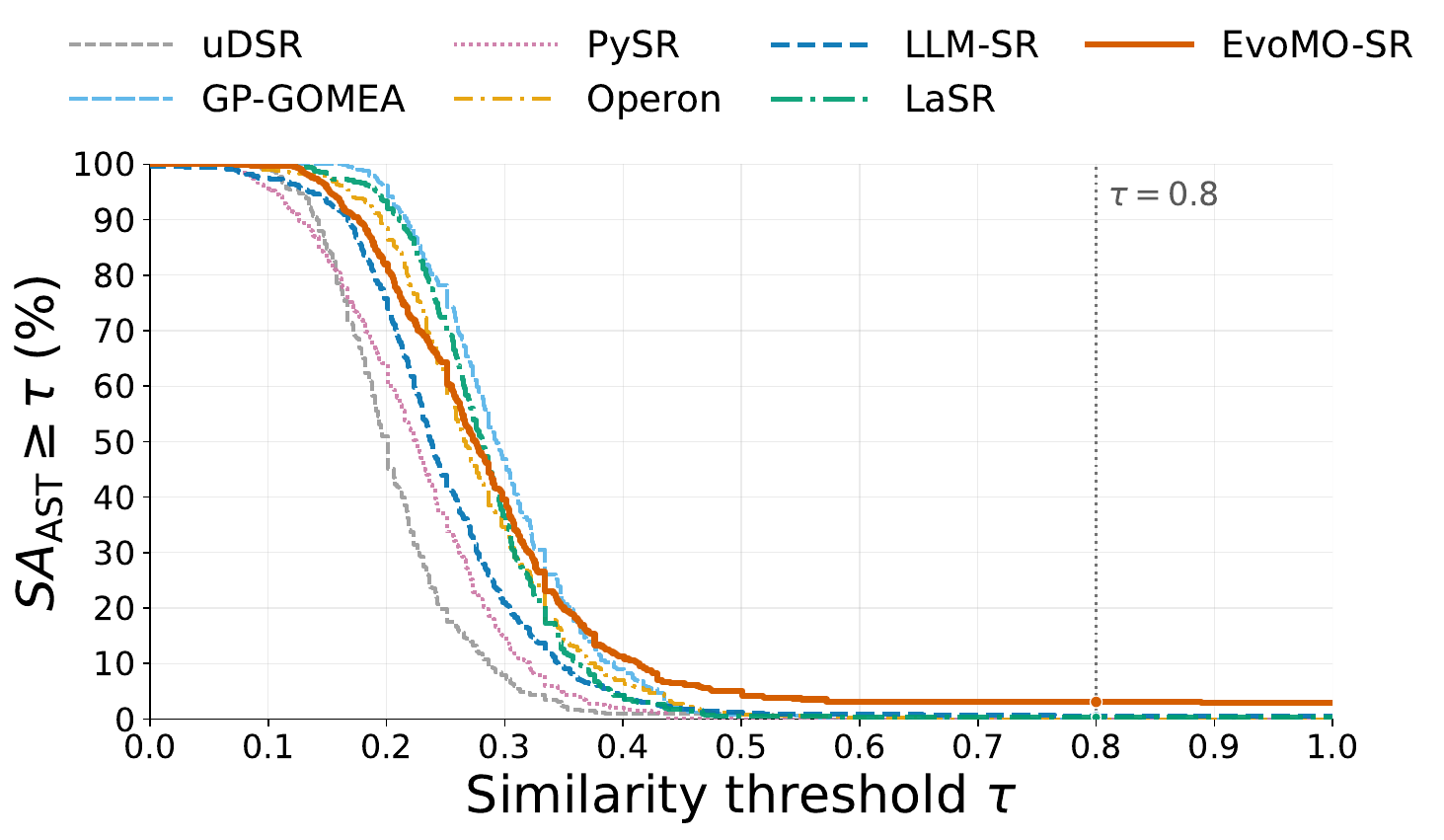}
        \caption{SA$_{\mathrm{AST}}$ threshold analysis.}
        \label{fig:sa_ast_threshold}
    \end{subfigure}
    %\hfill
    \begin{subfigure}[]{0.495\linewidth}
        \centering
        \includegraphics[width=\textwidth,trim=1mm 4mm 14mm 26mm,clip]{images/sa_term_threshold_recovery.pdf}
        \caption{SA$_{\mathrm{Term}}$ threshold analysis.}
        \label{fig:sa_term_threshold}
    \end{subfigure}
    \caption{Percentage of recovered expressions across all LSR-Synth datasets and runs (for a total of 645 cases) as a function of the threshold $\tau$ for \(SA_{AST}\) and \(SA_{Term}\). }
    \label{fig:sa_threshold_analysis}
\end{figure}

The improvements in numerical accuracy do not directly imply better symbolic recovery of the ground truth expressions. Figure \ref{fig:SA_lsr_phys} shows the average values of both SA metrics across all benchmarked methods over the LSR-Synth Physics dataset. The SA values are consistently low for every method, with GPGomea being the best for \(SA_{AST}\) and EvoMO-SR being the best for \(SA_{Term}\).
A more detailed analysis is presented in Figure \ref{fig:sa_threshold_analysis}, where we show the distributions of SA values according to a threshold \(\tau\) based on the results obtained over the LSR-Synth datasets (129 problems for 5 runs).
For \(SA_{AST}\) the curves are concentrated between roughly 0.15 and 0.4, and there is no single method dominating over this range. GP-GOMEA appears competitive, with one of the strongest curves around 0.2-0.3. EvoMO-SR becomes more dominant in the right tail, and around \(\tau \approx 0.35\) its curve becomes superior to the baselines, while the other methods collapse toward zero near \(\tau =1\).
For \(SA_{Term}\), LLM-SR is stronger than EvoMO-SR in retrieving similar expressions at very low thresholds. Around \(\tau \geq 0.3\), EvoMO-SR becomes the strongest method, with the advantage being clearer in the upper tail. 
These results indicate that EvoMO-SR's main advantage is not necessarily a uniform increase in structural similarity, but a greater probability of recovering highly accurate symbolic structures.

\subsection{Performance on Scientific Discovery Benchmarks}
We further evaluated EvoMO-SR on the scientific discovery benchmarks introduced by \cite{shojaee2025llmsr}. Among the scientific domains that the dataset covers, we choose the two physics datasets, i.e., \texttt{oscillator1} and \texttt{oscillator2}.  The training and ID samples are interleaved over \(t\in[30,50]\) (10'000 samples for the train set and 10'000 samples for the ID set), while OOD samples cover \(t\in[0,20]\) (10'000 samples), where \(t\) is the simulation time.
Results are presented in Table \ref{tab:otherdata}. EvoMO-SR outperforms both methods in accuracy achieving lower NMSE ID and OOD across 20 runs.
Concerning the symbolic similarity with the ground truth, LaSR obtained the best value for \(SA_{AST}\) in the \texttt{oscillator1} dataset, but its median \(SA_{Term}\) is 0 with around the 68\% of expressions having zero matching substructures with the ground truth. Even LLM-SR, despite achieving a higher SA value than EvoMO-SR on \texttt{oscillator2} and obtaining acceptable values for the other metric on \texttt{oscillator1} as well, fails to identify correct building blocks in the final expressions for \texttt{oscillator2}, resulting in a median \(SA_{Term}=0\) with the 60\% of expressions having no common additive terms with the ground truth.
To assess whether these differences are consistent across independent runs, we additionally performed paired Wilcoxon signed-rank tests between EvoMO-SR and LLM-SR. Detailed results are reported in Appendix \ref{app:stat_tests}. For NMSE, EvoMO-SR achieves lower values in the majority of paired runs, although the differences against LLM-SR do not reach statistical significance. The analysis over SA values shows a clearer distinction: \(SA_{AST}\) significantly favors LLM-SR on both datasets, while \(SA_{Term}\) significantly favors EvoMO-SR on \texttt{oscillator2}.
% The evaluation of SA values is shown in Figure \ref{fig:SA_others_osc}. The superiority of EvoMO-SR is more highlighted when comparing \(SA_{Term}\) values, meaning that our method was better for the identification of additive matching building blocks.

% \usepackage{booktabs}
% \usepackage{multirow}
\begin{table}[]
\centering
\caption{Direct comparison of median NMSE ID, NMSE OOD and SA values over 20 runs with LLM-SR and LaSR over  \texttt{oscillator1} and \texttt{oscillator2} datasets.}
\label{tab:otherdata}
\begin{tabular}{@{}ccccccccc@{}}

\multirow{2}{*}{\textbf{Method}} & \multicolumn{4}{c}{\textbf{\texttt{oscillator1}}}                                & \multicolumn{4}{c}{\textbf{\texttt{oscillator2}}}                                \\ \cmidrule(l){2-9} 
    & \textbf{ID} \(\downarrow\) & \textbf{OOD} \(\downarrow\) & \scriptsize\textbf{\(SA_{AST}\) \(\uparrow\)} & \scriptsize\textbf{\(SA_{Term}\) \(\uparrow\)} & \textbf{ID} \(\downarrow\) & \textbf{OOD} \(\downarrow\) & \scriptsize\textbf{\(SA_{AST}\) \(\uparrow\)} & \scriptsize\textbf{\(SA_{Term}\) \(\uparrow\)} \\ \midrule
LLM-SR                           & 1.55e-5          & 2.61e-2           & 0.19            & 0.28             & 2.85e-5          & 8.25e-3           & \textbf{0.23}   & 0                \\
LaSR                             & 4.37e-3          & 5.02e-1           & \textbf{0.27}   & 0                & 2.00e-1          & 1.15e-1           & 0.12            & 0                \\ \midrule
\textbf{EvoMO-SR}                & \textbf{1.53e-6} & \textbf{1.80e-2}  & 0.17            & \textbf{0.35}    & \textbf{1.69e-6} & \textbf{2.43e-4}  & 0.16            & \textbf{0.33}    \\
\end{tabular}
\end{table}
\subsection{Performance over randomly generated functions}
Although the datasets used before were designed to mitigate the phenomenon of memorization in LLMs, we still wanted to test our method on datasets generated by randomly constructed mathematical expressions. For each expression, we generated training and ID/OOD samples. Results in Table \ref{tab:random_results} show the comparison between our method and LLM-SR. The four ground truth expressions have an increasing degree of complexity, with the easiest (P1) retrieved by EvoMO-SR with median \(SA_{Term} = 1\).
\begin{table}[]
\centering
\caption{Median values of NMSE ID/OOD and SA metrics obtained by EvoMO-SR and LLM-SR across datasets generated with the following ground truth expressions: \(1.88\sin{x_1} + 1.61\cos(x_0)\) (P1), \(1.10x_0^3 + 2.02x_0^2-1.69x_1\) (P2), \(-1.95x_0^2-0.35x_1\sin{x_1}+2.36x_2^2+2.88\) (P3), \(1.90x_1^2+2.38x_1x_2-2.16\sin{x_0}-1.08\cos{x_0} -2.23\cos{x_2}-2.10\) (P4).}
\label{tab:random_results}
\begin{tabular}{@{}cccccc@{}}
\textbf{Problem} & \textbf{Method} & \textbf{ID} \(\downarrow\) & \textbf{OOD}  \(\downarrow\) & \(SA_{AST}\) \(\uparrow\) & \(SA_{Term}\) \(\uparrow\)\\ \midrule
\multirow{2}{*}{P1} & \textbf{EvoMO-SR} & \textbf{2.56e-32} & \textbf{1.40e-32} & \textbf{0.26} & \textbf{1.00} \\
 & LLM-SR & 2.49e-10 & 4.69e-8 & 0.19 & 0.14 \\ \midrule
\multirow{2}{*}{P2} & \textbf{EvoMO-SR} & \textbf{7.03e-15} & \textbf{1.94e-15} & 0.22 & \textbf{0.75} \\
 & LLM-SR & 5.89e-12 & 4.06e-11 & 0.22 & 0.37 \\ \midrule
\multirow{2}{*}{P3} & \textbf{EvoMO-SR} & \textbf{2.20e-7} & \textbf{4.02e-5} & 0.11 & 0.18 \\
 & LLM-SR & 2.06e-5 & 1.56e-3 & \textbf{0.20} & 0.18 \\ \midrule
\multirow{2}{*}{P4} & \textbf{EvoMO-SR} & \textbf{3.63e-6} & \textbf{1.66e-3} & 0.13 & \textbf{0.40} \\
 & LLM-SR & 1.23e-3 & 5.64e-2 & \textbf{0.19} & 0.17 \\ 
\end{tabular}
\end{table}

\subsection{Ablation studies}
\label{sec:ablation_studies}
The effectiveness of the substructure guidance mechanism has been evaluated by removing the substructure guidance module from the EvoMO-SR. To best test the differences in performance, we conducted this study using two larger models as well, namely DeepSeek-v4-Flash and GPT-5.4-nano. Table \ref{tab:llm_backbones} shows the results of this experiment over \texttt{oscillator1} and \texttt{oscillator2} datasets. The superiority of the substructure guidance approach is more highlighted for bigger models, especially for the \texttt{oscillator2} dataset with GPT-5.4-nano. Additional ablation studies are shown in Appendix \ref{app:add_abl}.

\begin{table}[]
\centering
\caption{Performance comparison across LLM backbones (Llama-3.1-8B-Instruct, DeepSeek-v4-Flash, GPT-5.4-nano) with substructure guidance (\textbf{with sub}) and without substructure guidance (\textbf{w/o sub}) over \texttt{oscillator1} and \texttt{oscillator2} datasets.}
\label{tab:llm_backbones}
\begin{tabular}{ccccccccc}
\multirow{3}{*}{\footnotesize \textbf{Model}} & \multicolumn{4}{c}{ \footnotesize \textbf{\texttt{oscillator1}}} & \multicolumn{4}{c}{ \footnotesize \textbf{\texttt{oscillator2}}} \\ \cline{2-9} 
 & \multicolumn{2}{c}{\footnotesize \textbf{w/o sub}} & \multicolumn{2}{c}{\footnotesize \textbf{with sub}} & \multicolumn{2}{c}{\footnotesize \textbf{w/o sub}} & \multicolumn{2}{c}{\footnotesize \textbf{with sub}} \\ \cline{2-9} 
 & \footnotesize \textbf{ID} \(\downarrow\) & \footnotesize \textbf{OOD} \(\downarrow\) & \footnotesize \textbf{ID} \(\downarrow\) & \footnotesize \textbf{OOD} \(\downarrow\) & \footnotesize \textbf{ID} \(\downarrow\) & \footnotesize \textbf{OOD} \(\downarrow\) & \footnotesize \textbf{ID} \(\downarrow\) & \footnotesize \textbf{OOD} \(\downarrow\) \\ \hline
\scriptsize Llama8B & 1.39e-5 & \textbf{2.06e-2} & \textbf{4.79e-6} & 2.11e-2 & 1.98e-5 & 3.83e-3 & \textbf{8.45e-6} & \textbf{1.60e-3} \\
\tiny DeepSeek-v4-Flash & 2.26e-6 & 2.98e-2 & \textbf{9.01e-7} & \textbf{1.32e-2} & 4.43e-4 & 1.88e-1 & \textbf{1.62e-4} & \textbf{2.76e-3} \\
\scriptsize GPT-5.4-nano & 5.34e-6 & 3.15e-2 & \textbf{9.13e-7} & \textbf{1.09e-2} & 1.28e-3 & 1.94e-2 & \textbf{8.37e-7} & \textbf{2.06e-4}
\end{tabular}
\end{table}

\section{Conclusions, Limitations, and Future Work}
We introduced EvoMO-SR, an LLM-driven SR framework that combines multi-objective survival selection with the extraction and reuse of symbolic substructures during evolution. On the LSR-Synth benchmark, EvoMO-SR achieves the lowest aggregate NMSE in seven of the eight domain-split comparisons, including all four OOD settings using a small model (i.e., LLama-3.1-8B-Instruct). The results obtained on the 4 datasets created from randomly generated expression show that the advantage persists even when the ground truth equations are not derived from known scientific models.
The analysis of structural similarity highlights the importance of distinguishing numerical accuracy from the recovery of the symbolic ground truth. EvoMO-SR improves term matching recovery in several settings and shows greater probabilities of recovering highly similar expressions, although its advantages are not uniform across similarity metrics or scientific domains.
The main limitations first concerns symbolic accuracy, which is consistently low even for EvoMO-SR, leaving much room for improvement in future developments. Furthermore, the substructure-guidance mechanism associates substructures with the performance of the related expression, which does not isolate their individual contribution. Future work will investigate more refined management of the substructure archive, such as more informative scores to the extracted substructures.

% The ablation studies reveal complementary but context-dependent effects of the framework’s components. Multi-objective selection consistently reduces expression size, whereas substructure guidance can improve predictive performance and term recovery while also increasing complexity. These observations motivate treating component reuse and complexity control as interacting design choices rather than independently beneficial additions.

\subsection*{AI use statement}

% (This section is \textbf{required} and does not count toward the page limit.)

% In this work, we used generative AI tools for [tasks with required disclosure].
% We have not used generative AI tools for [other tasks with required disclosure],
% and [the rest of the required disclosure tasks] are not applicable to this work.
% Additionally, we used generative AI tools for [tasks with recommended
% disclosure]. We have reviewed all AI-assisted work. [Elaborate. For example, “we
% checked LLM-generated research ideas for potential plagiarism through a manual
% literature survey”, “LLM-generated code was verified and tested for correctness
% by 2 authors”, etc.]. We take responsibility for the final content of this work,
% including text, claims or artifacts produced with the aid of generative AI.

% See the ICLR 2027 AI Policy for Authors for more details. This statement should
% not be more than 1 page.

% MINE
In this work, we used AI to revise the text and generate part of the code.
We have reviewed all AI-assisted work. 
We take responsibility for the final content of this work, including text, claims or artifacts produced with the aid of generative AI.

\subsection*{Ethics statement}
This work presents an LLM-driven method for generating symbolic expressions and has demonstrated good performance on datasets spanning various scientific domains (namely physics, chemistry, biology, and materials science). It is important to clarify that the equations proposed by EvoMO-SR should be considered hypotheses and not scientifically established solutions. The evaluation of the results obtained should always be accompanied by validation by domain experts.

\subsection*{Reproducibility statement}
The EvoMO-SR codebase is available at the following anonymous repository: \url{https://anonymous.4open.science/r/EvoMO-SR-47B6}.\\
Section \ref{sec:methodology} describes the methodology of the approach; Section \ref{sec:exp_results} describes the main experimental settings; Appendices \ref{app:add_ev_components} and \ref{app:prompts} report the evolutionary settings and details on the prompts, respectively.

\bibliographystyle{unsrt}  
\bibliography{references}

\appendix
\section{Implementation details}
\label{app:implementation_details}
\subsection{Parameters setting}
\label{app:add_ev_components}
EvoMO-SR generates symbolic expressions in a complete population-based evolutionary loop, with multiple customizable evolutionary parameters. The population size is regulated through the parameter \texttt{n\_parents}, while the number of offspring evolved at each generation is controlled by \texttt{n\_offspring}. In our implementation, both \texttt{n\_parents} and \texttt{n\_offspring} are equal to 10. Another important parameter is the number of generations, which establishes how many parent-offspring evolutions are performed. Given that \texttt{n\_offspring} is equal to 10 and the number of generations is set to 100, the total number of equations produced is 1'000. 
The expressions are evolved based on the evaluation of a fitness function, which in our case is the Normalized Mean Squared Error (NMSE). This fitness assigns scores at the very beginning of the algorithm (when the initial population is created) and then each time an offspring individual is generated. However, the selection of survival expressions is not solely based on accuracy. If \texttt{multi\_objective} is \texttt{True}, then the multi-objective survival mechanism is enabled. This module can be customized with any objectives. In our main implementation, the objectives are the NMSE and the count of math nodes, i.e., the accuracy and complexity, respectively. 
The parameter \texttt{elitism} decides which set of individuals undergoes Non-Dominated Sorting, and thus multi-objective survival selection. If it is enabled, then both parent and offspring are included for survival; otherwise survival considers only offspring. We choose \texttt{elitism=True} as it prevents good non-dominated parent to be excluded from the population. 
When the generation is over, the substructure archive is updated. Its default dimension is set to 10 and the guidance probability to \(p_{sub}=50\%\).
%, because, after several rounds of trials, it proved to be the best at retaining a sufficient number of promising substructures. The percentage of a substructure included in the mutation prompt was also tested, with 50\% proving to be the most convincing.
In this regard, the mutation prompts are fully customizable. In our main implementation, at each iteration (which corresponds to one mutation) it is randomly selected a mutation prompt among two: ``Refine the current expression while preserving useful terms.'' or ``Propose a new expression completely different from the current one.''. The individual chosen for mutation is selected through a parent selection strategy, regulated with an appropriate parameter. In this paper we choose \texttt{parent\_selection = "random"}, corresponding to the \textit{random with replacement} strategy. Other two possible strategies are \textit{tournament selection} and \textit{roulette}.
With a fixed rate (the 50\%), the mutation prompt is enriched with substructure guidance (see Appendix \ref{app:prompts} for additional details on the prompts used). 
Other parameters determine which substructures are suitable for inclusion and which are not. If a substructure is too complex and long, it is rejected because it deviates from being a building block and comes closer to being a complete expression. Similarly, substructures that are equivalent to individual constants or variables are excluded. In general, substructures are extracted from each parent-offspring pair by detecting additive terms that are present in the offspring expression and not in the parent expression. The expressions are first canonicalized with Sympy and products are expanded when necessary to expose hidden additive terms. The detected substructures are converted back to NumPy syntax before being considered for the archive. To avoid duplicates, each accepted substructure is mapped to a unique key in which learnable coefficients \texttt{c[i]} are replaced by a common placeholder and operands of commutative operations such as addition and multiplication are sorted.

The full implementation of EvoMO-SR includes other parameters typical of evolutionary search, though they are not relevant to this paper. These parameters are used to further customize the search, making our framework modular. For example, it is possible to include niching mechanisms, such as MAP-Elites, sharing, novelty, clearing. It is also possible to include an adaptive prompt mechanism, which consists in asking the LLM to change the current task prompt based on the previous output (although it was not effective in the SR task).

\subsection{Prompts}
\label{app:prompts}
The initial prompt \(\mathcal{P_{\text{init}}}\) is composed of the role specification, the task description, the scientific context and the format of the output. The specification of the role given to the LLM is contained within the task prompt, and explicitly tells: ``You are an expert specializing in Symbolic Regression". We decided to specify the role since it could be useful to better frame the SR task to the LLM. In fact, recent studies on incorporating a specific role into LLM prompts have shown improved performances \citep{wang2024role}, also on reasoning tasks \citep{kong2024better}.
The whole role+task prompt used is shown in Fig.\ref{fig:task-prompt}. This prompt is the same for all the experiments, except for the number of features that changes with the datasets used. An example of the scientific context is given in Fig. \ref{fig:scientific-prompt}. In our experiments, the contextual information is directly extracted from the metadata of the datasets used, namely LSR-Synth datasets \citep{shojaee2025llm} and \texttt{oscillator1} and \texttt{oscillator2} datasets \citep{shojaee2025llmsr}. Since LLM-SR is evaluated using these scientific priors, we decided to include them to ensure a fair comparison. 
Finally, Fig.\ref{fig:format-prompt} shows the format prompt, in which we specify the code structure of the solution and the general format of the response.

\begin{figure}[]
\centering

\begin{tikzpicture}[
    box/.style={
        text width=0.88\linewidth,
        align=left,
        inner xsep=5mm,
        inner ysep=4mm,
        line width=0.7pt,
        font=\ttfamily\scriptsize
    },
    prompt/.style={
        box,
        fill=yellow!15,
        draw=orange!55!black
    },
    response/.style={
        box,
        fill=green!10,
        draw=green!45!black
    }
]

\node[prompt] (p) {
\textbf{Task Prompt}\\[1mm]
You are an expert specializing in Symbolic Regression.\\
I have a dataset Z with 2 features (from Z[:,0] to Z[:,1]) and a target variable y. \\
Your task is to discover the mathematical function that maps these features to y. \\
Write a Python function named `discovered\_equation(Z,c)` where:
\begin{itemize}[label=-]
    \item `Z` is a 2D numpy array with shape (N,2) .
    \item `c` is a 1D numpy array of learnable constants.
\end{itemize}

Return ONLY the function using numpy (as np). Do not use features beyond index 1. \\
The optimization of the constants will be left to an external optimizer.  \\
CRITICAL: You must balance accuracy and complexity. Shorter, elegant equations that capture the underlying physics are strongly preferred over overly complex ones. 
};

\end{tikzpicture}

 \caption{The task prompt we use, which contains the specification of the role and the description of the equation discovery task.}
  \label{fig:task-prompt}
\end{figure}

% \begin{figure}[]
% \centering
%   \centering \includegraphics[width=0.8\linewidth]{images/task_prompt.png}
%  \caption{The task prompt we use, which contains the specification of the role and the description of the equation discovery task.}
%   \label{fig:task-prompt}
% \end{figure}

\begin{figure}[]
\centering

\begin{tikzpicture}[
    box/.style={
        text width=0.88\linewidth,
        align=left,
        inner xsep=5mm,
        inner ysep=4mm,
        line width=0.7pt,
        font=\ttfamily\scriptsize
    },
    prompt/.style={
        box,
        fill=yellow!15,
        draw=orange!55!black
    },
    response/.style={
        box,
        fill=green!10,
        draw=green!45!black
    }
]

\node[prompt] (p) {
\textbf{Scientific context}\\[1mm]
This is a Scientific Equation Discovery task. \\
Task description:\\
Find the mathematical function skeleton that represents acceleration in a damped nonlinear oscillator system with driving force, given data on position, and velocity.\\
Here is the symbolic meaning of the input variables:\\
Z[:,0] represents symbol 'x': current position \\
Z[:,1] represents symbol 'v': velocity
};

\end{tikzpicture}

 \caption{One example of scientific prompt. Here the context extracted from metadata of \texttt{oscillator1} dataset.}
  \label{fig:scientific-prompt}
\end{figure}

% \begin{figure}[]
% \centering
%   \centering \includegraphics[width=0.8\linewidth]{images/scientific_prompt.png}
%  \caption{One example of scientific prompt. Here the context extracted from metadata of \texttt{oscillator1} dataset.}
%   \label{fig:scientific-prompt}
% \end{figure}

\begin{figure}[]
\centering

\begin{tikzpicture}[
    box/.style={
        text width=0.88\linewidth,
        align=left,
        inner xsep=5mm,
        inner ysep=4mm,
        line width=0.7pt,
        font=\ttfamily\scriptsize
    },
    prompt/.style={
        box,
        fill=yellow!15,
        draw=orange!55!black
    },
    response/.style={
        box,
        fill=green!10,
        draw=green!45!black
    }
]

\node[prompt] (p) {
\textbf{Format prompt}\\[1mm]
Format example:\\
\hspace*{1.5em}\textasciigrave\textasciigrave\textasciigrave python\\
\hspace*{1.5em}import numpy as np\\[2mm]
\hspace*{1.5em}def discovered\_equation(Z, c):\\
\hspace*{3em}return HERE THE EQUATION\\
\hspace*{1.5em}\textasciigrave\textasciigrave\textasciigrave\\[4mm]

Give the response in the format:\\
\# Description: \textless short-description of the meaning of the analytical\\
relationships and why you gave this expression. Also include the term(s)\\
you have added, deleted and/or changed.\textgreater\\
\# Constants: \textless number of constants needed, e.g., 3\textgreater\\
\# Code:\\
\hspace*{1.5em}\textasciigrave\textasciigrave\textasciigrave python\\
\hspace*{3em}\textless code\textgreater\\
\hspace*{1.5em}\textasciigrave\textasciigrave\textasciigrave
};

\end{tikzpicture}

 \caption{The format prompt.}
  \label{fig:format-prompt}
\end{figure}

% \begin{figure}[]
% \centering
%   \centering \includegraphics[width=0.8\linewidth]{images/format_prompt.png}
%  \caption{The format prompt.}
%   \label{fig:format-prompt}
% \end{figure}

Once the initialization is done, the subsequent generations require an enriched prompt \(\mathcal{P}_{\text{loop}}\) in which we add the individual to evolve and one mutation prompt, which is randomly selected from a pre-fixed set, that can be customized as desired. In our settings, we have two possible mutation prompts that ask the LLM to refine the current expression or to propose a new one. Figure \ref{fig:example-refinement} shows an example of the LLM's response to the refinement option over a selected expression. The LLM gives the description of the mutated expression, explaining what has been changed. As discussed in the main paper, a substructure guidance mechanism is activated with a pre-defined rate. When enabled, the mutation prompt is enriched with a specific prompt that tells the LLM to include a substructure (selected from the archive of substructures) in addition to the mutation operation. Figure \ref{fig:mut-substructure} shows an example of LLM response to the mutation + substructure-guidance prompt. 

\begin{figure*}[]
\centering

\begin{tikzpicture}[
    promptbox/.style={
        draw=orange!55!black,
        fill=yellow!20,
        line width=0.8pt,
        text width=0.78\textwidth,
        align=left,
        inner xsep=7mm,
        inner ysep=5mm,
        font=\ttfamily\scriptsize
    },
    answerbox/.style={
        draw=green!45!black,
        fill=green!15,
        line width=0.8pt,
        text width=0.78\textwidth,
        align=left,
        inner xsep=7mm,
        inner ysep=5mm,
        font=\ttfamily\scriptsize
    }
]

% Top box
\node[promptbox, anchor=north west] (prompt) at (0,0) {%
\textbf{Prompt}\\[1mm]
The symbolic structure of the formula (to be evolved) is:\\
c[0] * Z[:, 0]\\[2mm]
Refine the current expression while preserving useful terms.%
};

% Bottom box
\node[
    answerbox,
    anchor=north west
] (answer) at ($(prompt.south west)+(0,-0.45cm)$) {%
\textbf{LLM response} \\[1mm]
\# Description: Refine the current expression by adding a linear function\\
of velocity to the product of a linear function of position and a\\
constant, capturing the essence of a damped nonlinear oscillator system\\
with driving force.\\
\# I have added a new term that uses a learnable constant to the product of\\
a linear function of position and a constant.\\
\# Constants: 2\\
\# Code:\\
\textasciigrave\textasciigrave\textasciigrave python\\
def discovered\_equation(Z, c):\\
\hspace*{2em}return c[0] * Z[:, 0] + c[1] * Z[:, 1]\\
\textasciigrave\textasciigrave\textasciigrave%
};

\end{tikzpicture}

  \caption{An example of the LLM response to the refinement option for one iteration of a run over the \texttt{oscillator1} dataset.}
  \label{fig:example-refinement}
\end{figure*}

% \begin{figure}[]
% \centering
%   \centering \includegraphics[width=0.8\linewidth]{images/example_refinement.png}

%   \caption{An example of the LLM response to the refinement option for one iteration of a run over the \texttt{oscillator1} dataset with Multi-objective DivEvoLLM-SR.}
%   \label{fig:example-refinement}
% \end{figure}

\begin{figure*}[]
\centering

\begin{tikzpicture}[
    promptbox/.style={
        draw=orange!55!black,
        fill=yellow!20,
        line width=0.8pt,
        text width=0.78\textwidth,
        align=left,
        inner xsep=7mm,
        inner ysep=5mm,
        font=\ttfamily\scriptsize
    },
    answerbox/.style={
        draw=green!45!black,
        fill=green!15,
        line width=0.8pt,
        text width=0.78\textwidth,
        align=left,
        inner xsep=7mm,
        inner ysep=5mm,
        font=\ttfamily\scriptsize
    }
]

% Top box
\node[promptbox, anchor=north west] (prompt) at (0,0) {%
\textbf{Mutation prompt + Substructure guidance}\\[1mm]
Refine the current expression while preserving useful terms.

\#\#\# SUBSTRUCTURE-GUIDED MUTATION \#\#\#\\
You must use the following archived substructure in the mutation: Z[:, 0]**3 (ONLY IF it is not already present in the expression to modify).\\
The returned full equation must still be a valid single-line numpy expression inside `discovered\_equation(Z, c)`.%
};

% Bottom box
\node[
    answerbox,
    anchor=north west
] (answer) at ($(prompt.south west)+(0,-0.45cm)$) {%
\textbf{LLM response} \\[1mm]
\# Description: The refined expression preserves the driving force term, the damping term, and the nonlinear stiffness term. I've replaced the exponential term with a power term for position, added a new term with a constant coefficient and a sinusoidal function of the velocity, and modified the coefficients of the existing terms.\\
\# Constants: 3\\
\# Code:\\
\textasciigrave\textasciigrave\textasciigrave python\\
import numpy as np \\
def discovered\_equation(Z, c):\\
\hspace*{2em} return c[0] * Z[:, 0] + c[1] * np.sin(Z[:, 1]) + c[2] * Z[:, 0]**3\\
\textasciigrave\textasciigrave\textasciigrave%
};

\end{tikzpicture}

  \caption{An example of the LLM response to the mutation prompt and substructure guidance for one iteration of a run over the \texttt{oscillator1} dataset.}
  \label{fig:mut-substructure}
\end{figure*}

\subsection{Detailed Definition of Symbolic Accuracy Metrics}
\label{app:symbolic_accuracy}

As discussed in Section \ref{sec:SA_metrics}, the evaluation of SA between the best expression found $\hat{f}^{\star}$ and the ground truth expression $f^{\star}$ is made with two metrics. 
The \(SA_{AST}\) metric measures structural similarity through the overlap between the AST subtrees of the two compared expressions, while the \(SA_{Term}\) metric evaluates structural similarity at the level of additive terms rather than complete AST subtrees. The second metric tends to penalize less the additive terms present in $\hat{f}^{\star}$ but not in $f^{\star}$, giving a more generous comparison.

To justify this claim, we show below an example taking the best expression obtained for the \texttt{oscillator1} problem and the values of \(SA_{AST}\) and \(SA_{Term}\).
Consider the discovered equation

\[
\hat{f}^{\star}(x,v) =
-0.1952\cdot x\exp(0.9942v)
+0.1408\cdot x^3
+0.8492\cdot vx-1.1545\cdot x\sin(v)
-0.5012\cdot v^3
\]

and the corresponding ground-truth expression

\[
f^{\star}(x,v)=
F\sin(\omega x)
-\alpha v^3
-\beta x^3
-\gamma xv
-x\cos(x).
\]

After the normalization employed by the term-based metric, their respective term sets become

\[
\mathcal{T}_{Term}(\hat{f}^{\star})= [
v^3,xv,x^3,x\exp(v),x\sin(v)]
\]

and

\[
\mathcal{T}_{Term}(f^{\star})= [
\sin(x),v^3,xv,x^3,x\cos(x)]
\]

The two expressions therefore share three normalized terms,

\[
\mathcal{T}_{Term}(\hat{f}^{\star})
\cap
\mathcal{T}_{Term}(f^{\star})=[v^3,xv,x^3]
\]

At the same time, the discovered expression contains two additional terms, $x\exp(v)$ and $x\sin(v)$, while two ground-truth terms, $\sin(x)$ and $x\cos(x)$, are missing. The union consequently contains seven distinct terms, yielding \(SA_{Term}=\frac{3}{7}
\approx 0.429\). For the same pair of expressions, the AST-based metric gives \(SA_{AST}
\approx 0.159.\)

The difference between the two SA values depends on the way in which they are computed. Indeed, for \(SA_{Term}\) the three correctly recovered terms \(v^3\), \(x^3\) and \(xv\) are directly recognized as matches, while each additional or missing term contributes once to the Jaccard union. On the other hand, for \(SA_{AST}\), non-matching building blocks introduce multiple distinct subtree signatures, thereby producing a stronger reduction in structural similarity.

These considerations are not intended to establish a preferred metric, but rather to highlight that different ways of calculating the symbolic similarity between expressions can lead to a reevaluation of the results obtained. While the first metric is very stringent, low values still indicate that the resulting expression contains noise that distances it from the ground truth. On the other hand, the second metric allows us to highlight common terms, and high values indicate that the approach has successfully recovered substructures present in the ground truth.

% \subsubsection{Normalization and canonicalization}
% Before the actual computation, the discovered and ground-truth expressions are normalized to reduce differences caused by alternative symbolic representations. The two metrics follow two slightly different pipelines to respect the distinct levels of granularity on which their computation is based.
% For \(SA_{AST}\), common syntactic variants are mapped to a uniform representation, with the normalization of function aliases and Numpy notation. A lightweight SymPy canonicalization is then applied to simplify powers and common factors.
% Subtraction and division are respectively rewritten as addition and multiplication by an inverse, and associative additions and multiplications are flattened. Commutative operands are also deterministically ordered. Finally, numerical coefficients are replaced by placeholders. 
% For \(SA_{Term}\), the expressions are similarly converted to a common symbolic syntax and parsed with SymPy. However, the normalization is designed to preserve and expose additive components. 
% Each term is then reduced to its variable-dependent structural core by removing multiplicative coefficients and other variable-independent factors. Thus, for example, $2x^3$ and $\beta x^3$ are both represented as $x^3$. 

\subsection{Random function datasets}
In the following, we briefly describe how we generated the datasets of randomly generated functions. The symbolic expressions were constructed by combining polynomial, trigonometric, and interaction terms. Each term was assigned a coefficient with a randomly chosen sign and a magnitude sampled uniformly between 0.35 and 3.0. Candidate expressions were filtered according to predefined complexity ranges and numerical validity criteria, ensuring that all input variables contributed to the target. Each training and ID set has 1'000 samples, using Latin hypercube sampling over \([-1,1]^d\), where \(d\) denotes the number of input variables, while the OOD set has 2'000 samples, each with exactly one input dimension sampled outside the ID range, i.e., from \([-2,-1.2]\cup[1.2,2]\), while the remaining dimensions stayed within \([-1,1]\). The extrapolated coordinate was balanced across input dimensions and randomly assigned to samples. Targets were computed directly from the generating expressions without additive noise.

\section{Symbolic Regression methods}
We compared EvoMO-SR against SR baselines, including two LLM-based methods. 
\paragraph{uDSR}
uDSR \citep{landajuela2022unified} is a modular hybrid framework for SR that extends DSR \citep{petersen2019deep}, unifying different strategies, including the integration of additional linear token and GP search in the decoding stage. In our comparisons, we set the batch size to \(1'000\) and the maximum iterations to \(100'000\).

\paragraph{PySR}
PySR \citep{cranmer2023interpretable} is an open-source symbolic regression framework designed for SR, based on a multi-population evolutionary algorithm. In our comparisons, we set the population size to \(20\) and the maximum number of evaluations to \(1'000\).

\paragraph{Operon}
Operon \citep{burlacu2020operon} is a high-performance framework for SR based on C++, which enhances GP with an efficient linear tree representation, low-memory evaluation, and the generation of offspring with a concurrency model. For our experiments, we set the population size to \(50 \), the maximum number of evaluations to \(1'000\), and the optimizer iterations to \(100\).
\paragraph{GPGomea}
GPGomea \citep{virgolin2021improving}
is a GP-based approach for SR that learns linkage information, i.e., dependencies between genotype positions, and uses Gene-pool Optimal Mixing to propagate useful expression patterns. For our experiments, we set the maximum number of evaluations to \(1'000\).

\paragraph{LLM-SR}
LLM-SR \citep{shojaee2025llmsr} performs SR using an LLM that evolves and iteratively refines expression skeletons, leaving the fitting of coefficients to off-the-shelf optimizers. To make a fair comparison with our framework, we set the number of generated functions to \(1'000\) and optimized constants with L-BFGS-B.

\paragraph{LaSR}
LaSR \citep{srwithlearnedconceptlibrary} does not use the LLM as the primary equation generator, but for the enhancement of the GP search performed via PySR. In our experiments, we set both the number of iterations and the number of cycles per iteration to 10, maintaining the population size to 33 with 15 total populations.

\section{Additional Results}
\label{app:add_results}

\subsection{Symbolic Accuracy}
\label{app:add_results_SA}

\begin{figure}
    \centering
    \includegraphics[width=0.8\linewidth]{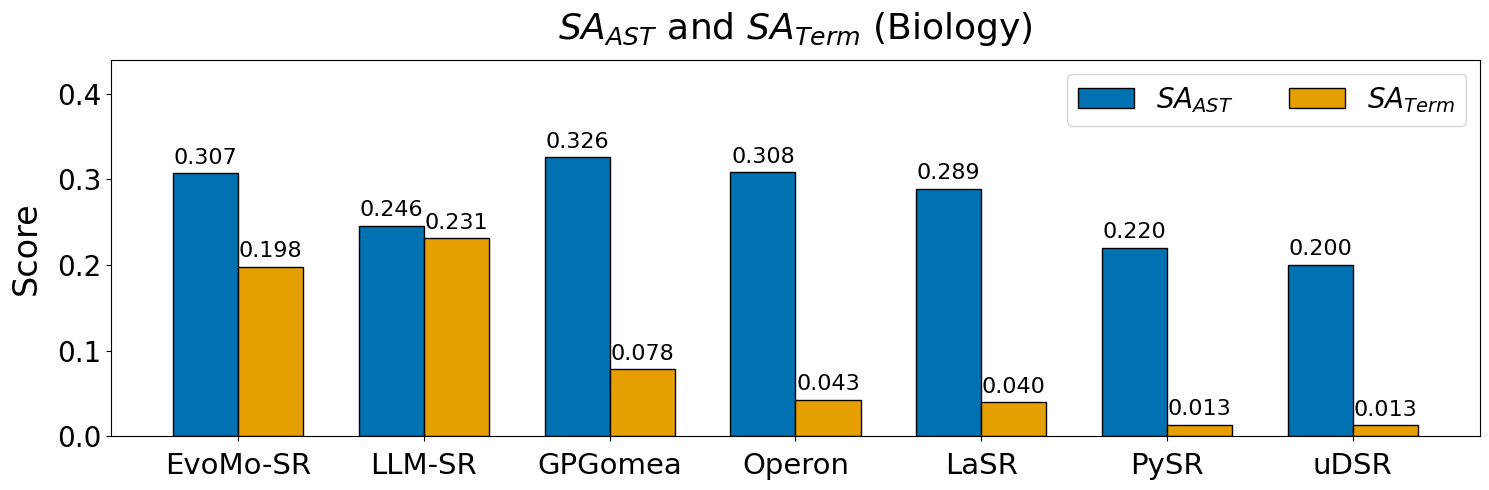}
    \caption{Comparison of \(SA_{AST}\) and \(SA_{Term}\) between EvoMO-SR and the other SR baselines over LSR-Synth Biology dataset.}
    \label{fig:SA_lsr_bio}
\end{figure}

\begin{figure}
    \centering
    \includegraphics[width=0.8\linewidth]{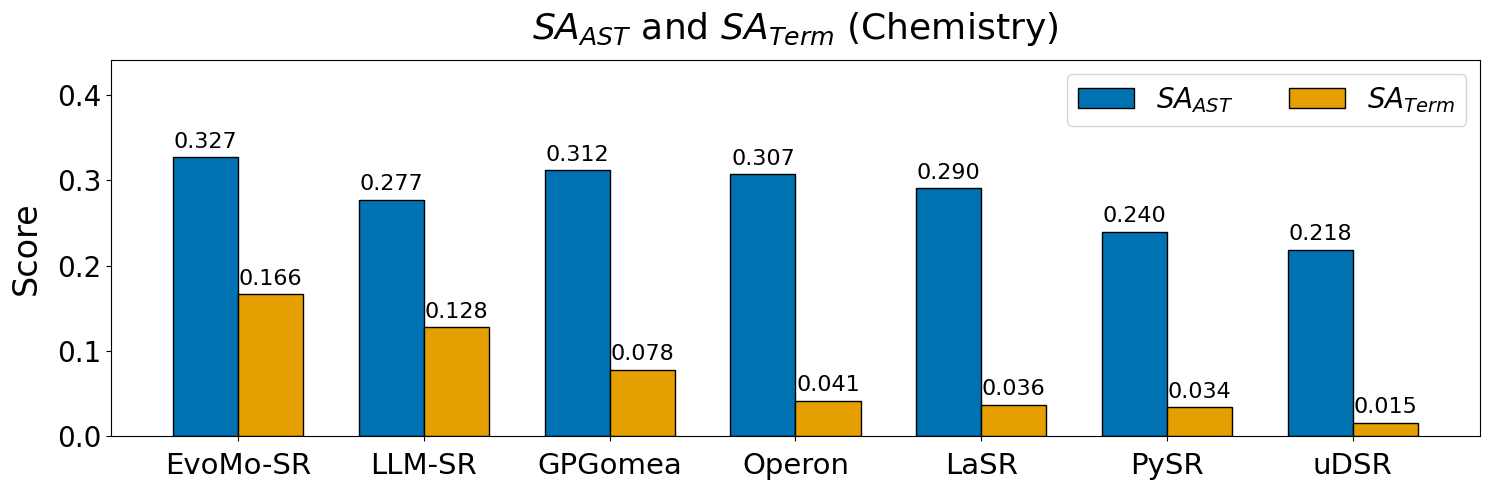}
    \caption{Comparison of \(SA_{AST}\) and \(SA_{Term}\) between EvoMO-SR and the other SR baselines over LSR-Synth Chemistry dataset.}
    \label{fig:SA_lsr_chem_react}
\end{figure}

\begin{figure}
    \centering
    \includegraphics[width=0.8\linewidth]{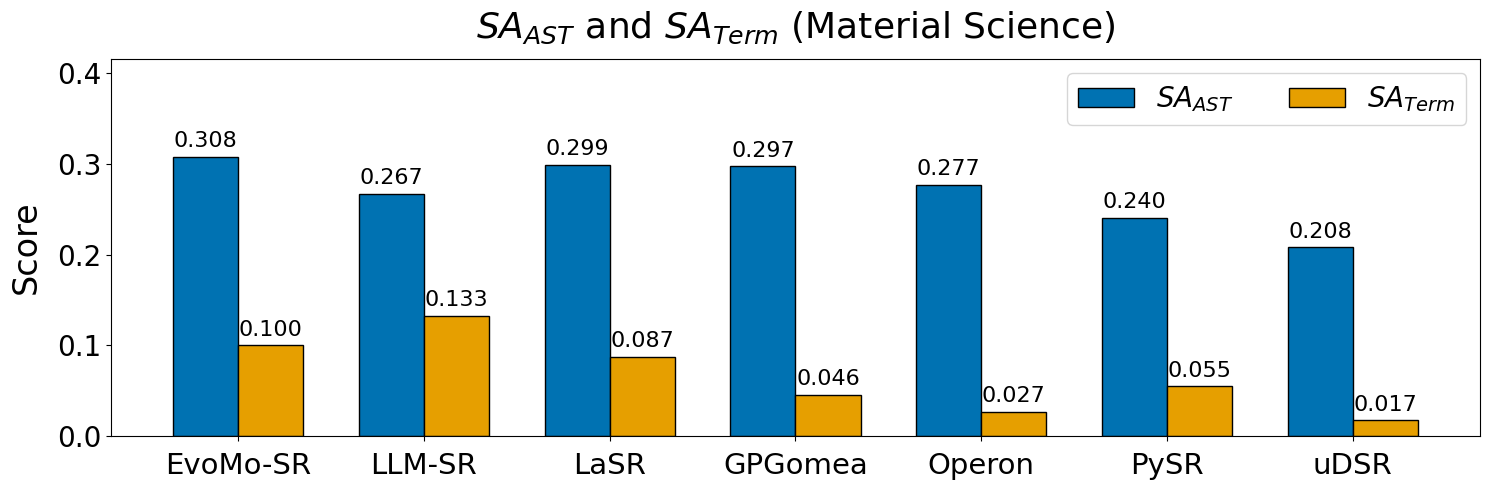}
    \caption{Comparison of \(SA_{AST}\) and \(SA_{Term}\) between EvoMO-SR and the other SR baselines over LSR-Synth Material Science dataset.}
    \label{fig:SA_lsr_matsci}
\end{figure}

In Section \ref{sec:lsr-synth-results} are shown the results of \(SA_{AST}\) and \(SA_{Term}\) of EvoMO-SR, LLM-SR, LaSR and the other GP-based SR baselines over the LSR-Synth Physics dataset. For the sake of completeness, we report here the results over the other three dataset: Biology (see Fig. \ref{fig:SA_lsr_bio}), Chemistry (see Fig. \ref{fig:SA_lsr_chem_react}), and Material Science (see Fig. \ref{fig:SA_lsr_matsci}).
EvoMO-SR always obtains better results on \(SA_{AST}\) compared to the two main competitors, i.e., LLM-SR and LaSR, except for the Physics domain (see Figure \ref{fig:SA_lsr_phys}) where LaSR is slightly superior. However, it is not always the best compared to GP-based baselines. Indeed, for the Biology and Physics dataset, GPGomea reached \(SA_{AST}=0.326\) and \(SA_{AST}=0.277\) respectively, which are the highest value obtained across all the methods. This is not surprising as GPGomea has already showed its goodness in symbolic recovery \citep{de2024srbench++}.
Concerning \(SA_{Term}\), EvoMO-SR wins in the Chemistry and Physics datasets, with LLM-SR being the hardest competitor. 

\subsection{Examples of expressions obtained}
Below are some of the best results from the experiments using LLM-driven methods. 

\subsubsection{\texttt{oscillator1} dataset}
The fitted ground truth for the \texttt{oscillator1} dataset is the following expression:

\[
\dot{v} = 
0.8\sin{x} -0.5 v^3 -0.2 x^3 -0.5 xv -x\cos{x}.
\]

where \(v\) is the velocity and \(x\) the position.

The best expression obtained for \texttt{oscillator1} for LLM-SR achieved NMSE ID and OOD equal to 1.87e-11 and 2.46e-10, respectively, and is:
\[\dot{v} = \text{gradient}(v,0.002)\]

Basically, LLM-SR retrieved a formula which numerically differentiates the velocity vector through the \texttt{gradient} operator provided by NumPy. Since acceleration is itself the time derivative of velocity and the observations are sampled at a fixed temporal resolution, this expression provides an extremely accurate approximation of the target, explaining why NMSE ID and OOD are so low. However, for the \texttt{oscillator1} datasets, this shortcut is taken only once out of 20 runs. 
For example, the second best expression among the 20 runs based on accuracy is:
\[\dot{v} = -0.4648x + 0.0008v -0.00022\sin{(2\pi\cdot 1.4157x)} + \]
\[+ 0.1495x^3 -0.4998xv + 0.2688v^2 -0.5228v^3 + 0.2664(x - v^2)\]
which obtains NMSE ID and OOD of 6.06e-7 and 2.33e-2, respectively. This expression was also the one with highest \(SA_{Term}\). Here, the symbolic expression is much closer to the ground truth, sharing some additive terms such as \(x^3\) and \(v^3\).

The best expression among the 20 runs based on accuracy for EvoMO-SR was 
\[\dot{v} = 0.0889\cos{(2.2528x +1.5709)} - 0.4999xv - 0.4997v^3\]
and gives NMSE ID and OOD of 6.35e-8 and 1.78e-4, respectively.
% \[\dot{v} = 0.2999sin(t) -2.1033x -0.001v -0.8877x^2  + -9.4404e^{-06} \cdot exp(v)\cdot cos(t)+\]
% \[ -1.6080x\cdot exp(t) -0.4983v^3-1.000xv -1.2894 * sin(x)\]
% and gives NMSE ID and OOD of 1.10e-8 and 2.58e-6. respectively. 

The one with the highest \(SA_{Term}\) (equal to \(0.57\)) is:

\[\dot{v} = -0.0147 -0.0463x^3 +0.0004\exp{(-x)} + 1.0216x -0.4994v^3+\]
\[-0.4999xv +0.4882(-2.5\sin{x} + 0.0293) \]
and gives NMSE ID and OOD of 8.66e-7 and 1.54e-2. respectively.

\subsubsection{\texttt{oscillator2} dataset}
The ground truth for the \texttt{oscillator2} dataset is:
\[ \dot{v} = 0.3\sin{t} - 0.5v^3 -xv -5.0x\cdot \exp{(0.5x)}\]
For the \texttt{oscillator2} dataset, LLM-SR returned expressions involving the \texttt{gradient} operator in 6 out of 20 runs, showing that the use of this operator might represent a recurring behavior.
In the run that achieves the lowest NMSE ID and OOD (9.74e-10 and 1.38e-10, respectively), the expression is 
\[ \dot{v} = \text{gradient}(-1.48e^{-6}t -2.57e^{-6}x+v+0.99,1)\]
while the second and third best expressions (always according to NMSE) are both equal to \(\text{gradient}(v,t)\). So, similarly to the case of the first dataset, also here the method tries to employ a shortcut to approximate the acceleration. 
EvoMO-SR never proposes final expressions with non-conventional operators such as the \texttt{gradient}. The expression with the lowest NMSE ID and OOD (1.10e-8 and 2.58e-6, respectively) is also the one with the highest \(SA_{Term}\) (equal to \(0.44\)), and is the following:
\[ \dot{v} = 0.2999\sin{t} -2.1033x -0.0001v -0.8877x^2 +\]
\[-9.4404e^{-6}\exp{v} *\cos{t} -1.6080x\cdot \exp{x} -0.4984v^3-1.0003xv -1.2894\sin{x}\]

\subsection{Additional ablation studies}
\label{app:add_abl}
We performed additional ablation studies by alternatively removing the main components of EvoMO-SR: substructure guidance and multi-objective selection. We carried out 3 different experiments: without substructure guidance and without multi-objective selection (i.e., selection is performed based on accuracy only), without substructure guidance and with multi-objective selection, with substructure guidance and without multi-objective selection. The results of these experiments (each configuration was run 5 times with Llama-3.1-8B-Instruct as the LLM backbone) are shown in Table \ref{tab:abl-osc1-compr} (for the \texttt{oscillator1} dataset) and Table \ref{tab:abl-osc2-compr} (for the \texttt{oscillator2} dataset), which compare NMSE ID and OOD, the two similarity metrics and the average complexity (measured counting the number of nodes in the final best expressions). As expected, the most consistent effect of multi-objective selection was a reduction in expression size for both dataset and both configuration with multi-objective selection enabled. This aspect is crucial because reduced complexity leads to greater interpretability of the resulting expressions, and this becomes even more important when this trade-off does not reduce overall accuracy. 
The effect of substructure guidance has to be analyzed carefully. For the \texttt{oscillator1} dataset, as already observed in Section \ref{sec:ablation_studies}, the use of substructure guidance slightly improves NMSE ID compared to the configuration without guidance and with multi-objective selection. The improvement in symbolic accuracy can be seen more for \(SA_{Term}\), while the complexity increases. For the \texttt{oscillator2} dataset, substructure guidance in EvoMO-SR improved symbolic accuracy values and NMSE OOD against all the configurations.
Overall, we can state that substructure guidance did not consistently improve the metrics when considered as an isolated main effect, suggesting an interaction with multi-objective selection rather than a uniform benefit.

\begin{table}[]
\centering
\caption{Comprehensive ablation study removing substructure guidance (Sub) and multi-objective (MO) selection mechanism over \texttt{oscillator1} dataset.}
\label{tab:abl-osc1-compr}
\begin{tabular}{clccccc}
\multicolumn{2}{c}{\textbf{Method}} & \multicolumn{5}{c}{\textbf{Oscillator 1}} \\ \hline
\textbf{Sub} & \textbf{MO} & \textbf{ID} \(\downarrow\) & \textbf{OOD} \(\downarrow\) & \textbf{\(SA_{AST}\) \(\uparrow\)} & \textbf{\(SA_{Term}\) \(\uparrow\)} & \textbf{Complexity} \(\downarrow\) \\ \hline
\(\times\) & \(\times\) & 4.29e-6 & 2.30e-2 & 0.14 & 0.27 & 121.6 \\
\(\times\) & \(\checkmark\) & 1.39e-5 & 2.06e-2 & \textbf{0.31} & 0.37 & \textbf{60.2} \\
\(\checkmark\) & \(\times\) & \textbf{3.81e-6} & \textbf{1.70e-2} & 0.15 & 0.32 & 112.4 \\ \hline
\textbf{\(\checkmark\)} & \textbf{\(\checkmark\)} & 4.79e-6 & 2.11e-2 & 0.28 & \textbf{0.41} & 86.8
\end{tabular}
\end{table}

\begin{table}[]
\centering
\caption{Comprehensive ablation study removing substructure guidance (Sub) and multi-objective (MO) selection mechanism over \texttt{oscillator2} dataset.}
\label{tab:abl-osc2-compr}
\begin{tabular}{clccccc}
\multicolumn{2}{c}{\textbf{Method}} & \multicolumn{5}{c}{\textbf{Oscillator 2}} \\ \hline
\textbf{Sub} & \textbf{MO} & \textbf{ID} \(\downarrow\) & \textbf{OOD} \(\downarrow\) & \textbf{\(SA_{AST}\) \(\uparrow\)} & \textbf{\(SA_{Term}\) \(\uparrow\)} & \textbf{Complexity} \(\downarrow\) \\ \hline
\(\times\) & \(\times\) & \textbf{1.38e-6} & 1.34e-2 & 0.14 & 0.27 & 171.0 \\
\(\times\) & \(\checkmark\) & 1.98e-5 & 3.83e-3 & 0.29 & 0.32 & \textbf{94.0} \\
\(\checkmark\) & \(\times\) & 5.03e-6 & 2.69e-3 & 0.11 & 0.16 & 214.8 \\ \hline
\textbf{\(\checkmark\)} & \textbf{\(\checkmark\)} & 8.45e-6 & \textbf{1.60e-3} & \textbf{0.29} & \textbf{0.35} & 105.6
\end{tabular}
\end{table}

\subsubsection{GP-based mutation operator}
\begin{figure}
    \centering
    \includegraphics[width=1.0\linewidth]{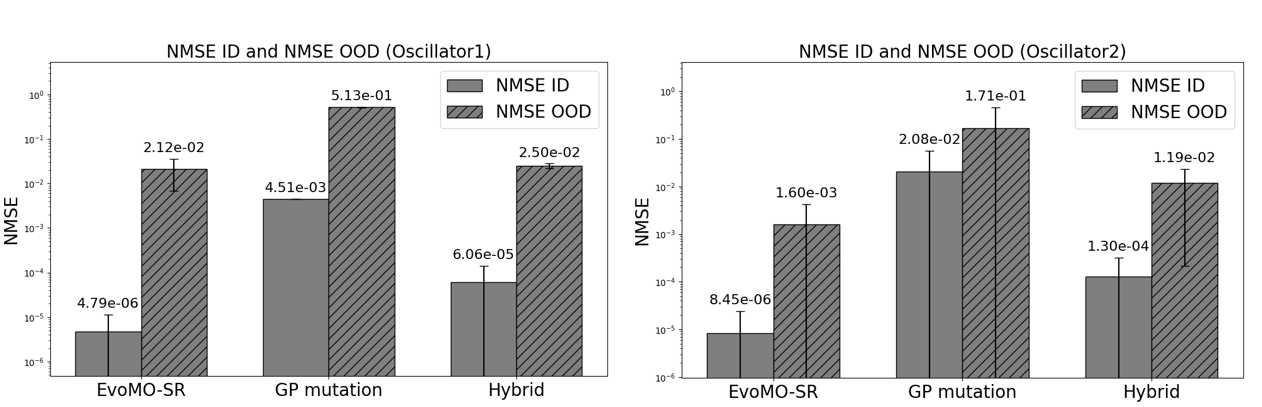}
    \caption{Comparison of NMSE ID and OOD values between EvoMO-SR with LLM-driven mutation (original), with GP mutation, and hybrid mutations (both LLM-driven and GP-based).}
    \label{fig:GP_mutation}
\end{figure}

Another ablation study concerned the use of LLM for the mutation of individuals. In EvoMO-SR, the LLM mutates the expressions with the prompt ``Refine the current expression while preserving useful terms''. There is no control over the result of this mutation: the LLM can deliberately delete one or more terms, add one or more terms, modify existing substructures etc. Since the LLM is integrated in an evolutionary loop, it is easy to replace the LLM-driven mutation with a standard GP mutation operator (i.e., \texttt{mutUniform}, which randomly select a subtree in the expression tree with another randomly generated subtree) in order to analyze whether the mutation strategy has impact on the results. Figure \ref{fig:GP_mutation} shows the average NMSE ID and OOD values obtained in 5 runs over the \texttt{oscillator1} and \texttt{oscillator2} datasets by replacing LLM mutation with traditional GP mutation and with an hybrid approach (i.e., LLM mutation alternated with GP mutation). The mutation prompt concerning the proposal of a new expression was disabled for this experiment. This ablation study clearly shows better accuracy performance in favour of EvoMO-SR (with LLM mutation). 

\subsection{Statistical tests}
\label{app:stat_tests}
In Section \ref{sec:lsr-synth-results} we showed the comparison of NMSE ID and OOD values between EvoMO-SR and the other baselines across LSR-Synth datasets. The results highlight the superiority of our method in almost all domains, except for Physics, where LLM-SR outperforms it. To statistically demonstrate this superiority, we performed a Wilcoxon test with Holm correction between EvoMO-SR and its main competitor, LLM-SR, using each benchmark problem as the paired experimental unit. For each problem, NMSE was first averaged across the five independent runs. We considered ID and OOD performance separately and applied Holm correction to account for multiple comparisons. Table \ref{tab:test_evo_llmsr} reports the resulting Holm-adjusted $p$-values together with the number of problems on which each method achieves lower NMSE. 
% \begin{table}[]
% \centering
% \caption{Domain-specific pairwise Wilcoxon test with Holm's correction between EvoMO-SR and LLM-SR}
% \label{tab:test_evo_llmsr}
% \begin{tabular}{@{}lll@{}}
% \toprule
% \textbf{Domain}  & \textbf{Split} & \textbf{p-value} \\ \midrule
% Physics          & ID             & 0.885            \\
% Physics          & OOD            & 0.007            \\
% Chemistry        & ID             & 0.001            \\
% Chemistry        & OOD            & 3.33e-5          \\
% Biology          & ID             & 3.00e-5          \\
% Biology          & OOD            & 2.00e-5          \\
% Material Science & ID             & 8.00e-6          \\
% Material Science & OOD            & 5.00e-6          \\ \bottomrule
% \end{tabular}
% \end{table}

\begin{table}[]
\caption{Domain-specific pairwise Wilcoxon test with Holm's correction between EvoMO-SR and LLM-SR}
\centering
\label{tab:test_evo_llmsr}
\begin{tabular}{@{}cccccc@{}}

\textbf{Domain}                   & \textbf{Split} & \textbf{p-value} & \textbf{EvoMO-SR (ours) wins} & \textbf{LLM-SR wins} & \textbf{Ties} \\ \midrule
\multirow{2}{*}{Physics}          & ID             & 0.885            & \textbf{24}            & 20                   & 0             \\
    & OOD            & 0.007            & \textbf{30}            & 14 & 0             \\ \midrule
\multirow{2}{*}{Chemistry}        & ID             & 0.001            & \textbf{25}            & 11                   & 0             \\
        & OOD            & 3.33e-5          & \textbf{31}            & 5                    & 0             \\ \midrule
\multirow{2}{*}{Biology}          & ID             & 3.00e-5          & \textbf{22}            & 2                    & 0             \\
     & OOD            & 2.00e-5          & \textbf{22}            & 2                    & 0             \\ \midrule
\multirow{2}{*}{Material Science} & ID             & 8.00e-6          & \textbf{24}            & 1                    & 0             \\
     & OOD            & 5.00e-6          & \textbf{24}            & 1                    & 0             \\ 
\end{tabular}
\end{table}

These results confirm that the differences generally favor EvoMO-SR. For Chemistry, Biology, and Material Science, EvoMO-SR achieves significantly lower NMSE than LLM-SR for both ID and OOD evaluation after Holm's correction ($p < 0.05$). For the Physics dataset, no statistically significant difference is observed for ID performance ($p=0.885$). However, EvoMO-SR significantly outperforms LLM-SR on Physics OOD data winning on 30 out of 44 problems. Overall, EvoMO-SR consistently reaches lower NMSE ID and OOD with statistical significance, obtaining 95 wins out of 129 problems for ID performance and 107 wins out of 129 problems for OOD performance. 

The same statistical test was carried out for the results obtained over \texttt{oscillator1} and \texttt{oscillator2} datasets. Table \ref{tab:test_evo_llmsr_other_data} shows results obtained performing the Wilcoxon paired test over NMSE ID and OOD values. Even though EvoMO-SR always achieves more wins against LLM-SR, the p-value scores are above 0.05, meaning that the test does not show any statistical significance.

The two methods were also compared on the SA values, as shown in Table \ref{tab:test_evo_llmsr_other_data_SA}. Here, almost all the results are statistically significant, obtaining a p-value score below 0.05, except for \(SA_{Term}\) in the \texttt{oscillator1} datasets.
\begin{table}[]
\centering
\caption{Pairwise Wilcoxon test with Holm's correction between EvoMO-SR and LLM-SR over \texttt{oscillator1} and \texttt{oscillator2} datasets}
\label{tab:test_evo_llmsr_other_data}
\begin{tabular}{@{}cccccc@{}}

\textbf{Domain}                & \textbf{Split} & \textbf{p-value} & \textbf{EvoMO-SR (ours) wins} & \textbf{LLM-SR wins} & \textbf{Ties} \\ \midrule
\multirow{2}{*}{\texttt{oscillator1}} & ID             & 0.09             & \textbf{14}            & 6                    & 0             \\
     & OOD             & 0.24             & \textbf{12}            & 8                    & 0             \\ \midrule
\multirow{2}{*}{\texttt{oscillator2}} & ID             & 0.09             & \textbf{13}            & 7                    & 0             \\
       & OOD       & 0.09          & \textbf{13}            & 7                    & 0 
\end{tabular}
\end{table}

\begin{table}[]
\centering
\caption{Pairwise Wilcoxon test with Holm's correction over SA metrics between EvoMO-SR and LLM-SR over \texttt{oscillator1} and \texttt{oscillator2} datasets.}
\label{tab:test_evo_llmsr_other_data_SA}
\begin{tabular}{@{}cccccc@{}}

\textbf{Domain}                & \textbf{Metric} & \textbf{p-value} & \textbf{EvoMO-SR (ours) wins} & \textbf{LLM-SR wins} & \textbf{Ties} \\ \midrule
\multirow{2}{*}{\texttt{oscillator1}} & \(SA_{AST}\)         & 0.01             & 5                      & \textbf{15}          & 0             \\
    & \(SA_{Term}\)    & 0.13             & \textbf{13}            & 6                    & 1             \\ \midrule
\multirow{2}{*}{\texttt{oscillator2}} & \(SA_{AST}\)         & 0.001            & 3                      & \textbf{17}          & 0             \\
         & \(SA_{Term}\)      & 0.001            & \textbf{17}            & 1                    & 2             \\ 
\end{tabular}
\end{table}

% \begin{table}[]
% \centering
% \caption{Pairwise Wilcoxon test between EvoMO-SR and LLM-SR for \texttt{oscillator1} and \texttt{oscillator2} datasets}
% \label{tab:test_evo_llmsr_other_data}
% \begin{tabular}{@{}cccccc@{}}
% \toprule
% \textbf{Dataset} & \textbf{Split} & \textbf{p-value} & \textbf{EvoMO-SR wins} & \textbf{LLM-SR wins} & \textbf{Ties} \\ \midrule
% \texttt{oscillator1}     & ID             & 0.062            & 5                      & 0                    & 0             \\
% \texttt{oscillator1}      & OOD            & 0.625            & 4                      & 1                    & 0             \\
% \texttt{oscillator2}       & ID             & 0.062            & 0                      & 5                    & 0             \\
% \texttt{oscillator2}       & OOD            & 0.062            & 0                      & 5                    & 0            
% \end{tabular}
% \end{table}

\section{Pareto fronts analysis}
\label{app:pareto_fronts}

\begin{figure}
    \centering
    \includegraphics[width=1.0\linewidth]{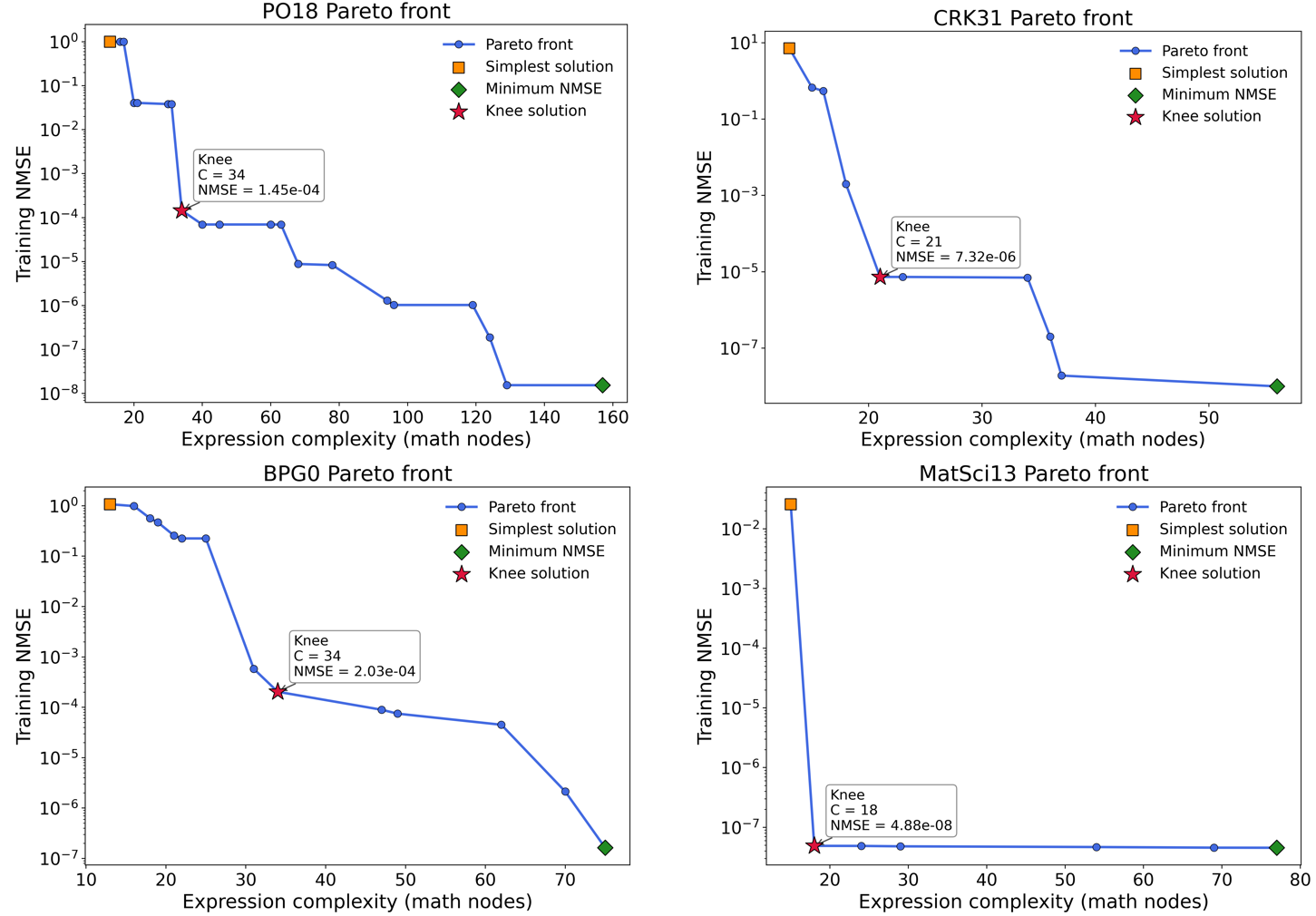}
    \caption{Obtained approximated Pareto fronts of distinct runs over 4 problems: PO18 for the Physics domain, CRK31 for the Chemistry domain, BPG0 for the Biology domain, and MatSci13 for the Material Science domain.}
    \label{fig:pareto_fronts}
\end{figure}

Figure \ref{fig:pareto_fronts} shows 4 distinct final approximated Pareto fronts each obtained from a run of an individual problem. Each Pareto front is computed with the Non-Dominated Sorting algorithm balancing accuracy (i.e., NMSE) and complexity (count of math nodes). From the final Pareto front is possible to extract the best expression according to a predefined criteria, which can be the knee point, the one with the lowest complexity and the one with the lowest NMSE (highest accuracy). However, the choice is not always straightforward. It is clear that the geometry of the Pareto front changes with the problem. Only for the MatSci13 problem the knee point seems the best alternative, as an increase in complexity does not improve the accuracy significantly. For all the other three cases, an increase in complexity beyond the knee always brings an increase in accuracy by several orders of magnitude. 

\begin{figure}[b]
    \centering
    \includegraphics[width=1.0\linewidth]{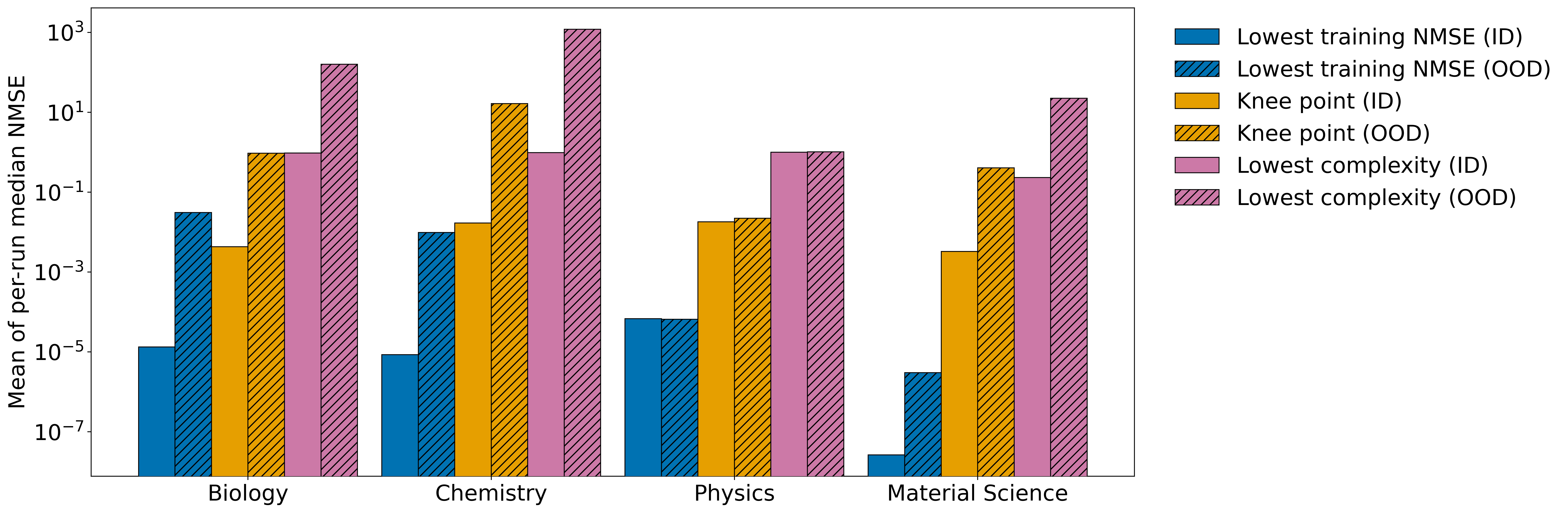}
    \caption{Average per-run Median NMSE ID and OOD values of the best expressions selected based on the lowest training NMSE, the lowest complexity and the knee point of the final pareto front.}
    \label{fig:pareto_front_NMSE}
\end{figure}

\begin{figure}[t]
    \centering
    \includegraphics[width=0.6\linewidth]{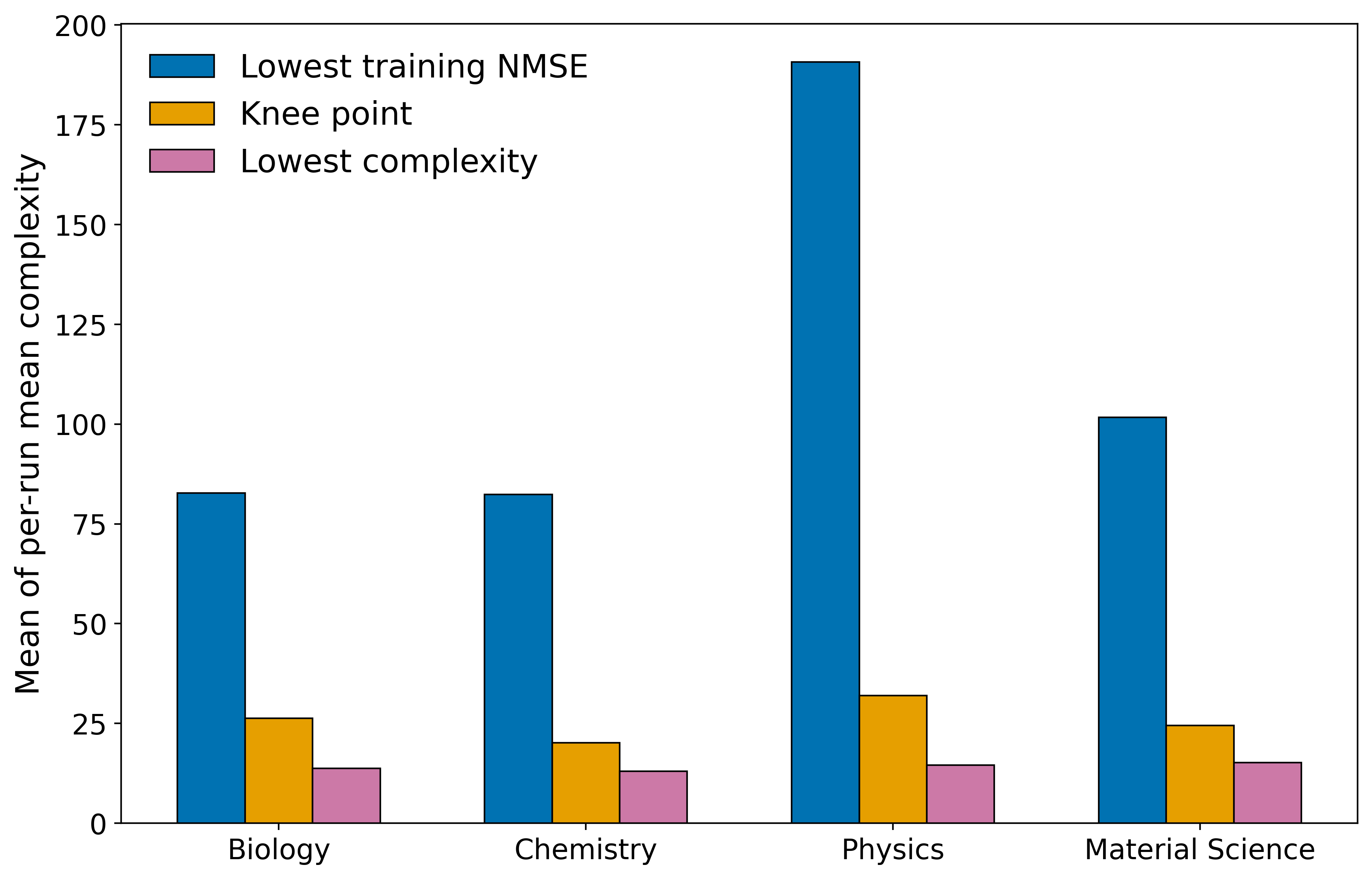}
    \caption{Average per-run Mean complexity values of the best expressions selected based on the lowest training NMSE, the lowest complexity and the knee point of the final pareto front.}
    \label{fig:pareto_front_complexity}
\end{figure}

\begin{figure}[t]
    \centering
    \includegraphics[width=1.0\linewidth]{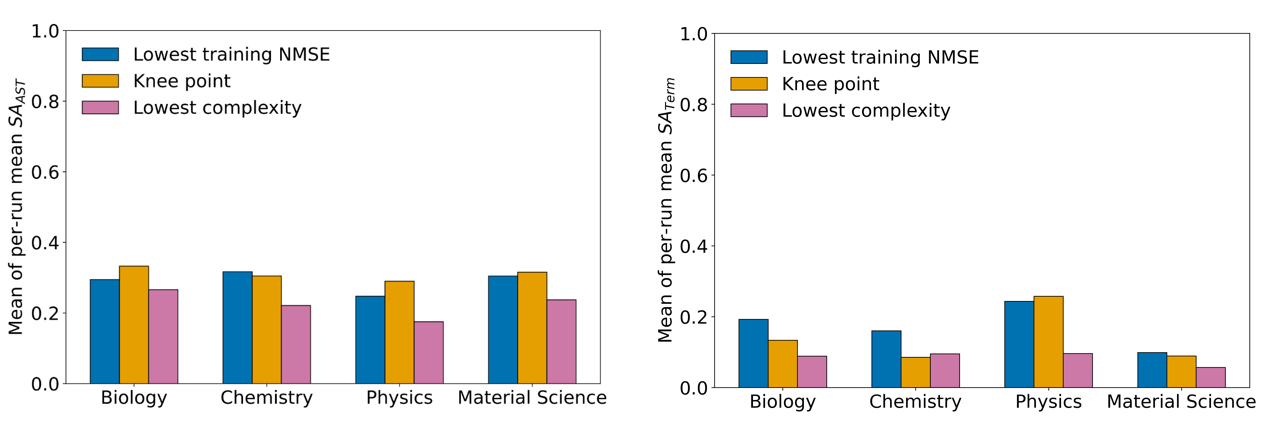}
    \caption{Average per-run Mean similarity values of the best expressions selected based on the lowest training NMSE, the lowest complexity and the knee point of the final pareto front. On the left results concerning \(SA_{AST}\) values, while on the right results concerning \(SA_{Term}\) values.}
    \label{fig:pareto_front_similarity}
\end{figure}

The results reported in this paper are obtained by selecting the best expression based on minimum training NMSE, as the main preference was to favour more accurate expressions even if more complex. To justify this choice, we studied the NMSE ID and OOD, the complexity and the \(SA\) values comparing best expressions selected from the final Pareto front according to lowest training NMSE, knee point and lowest complexity. Figure \ref{fig:pareto_front_NMSE} shows that, as expected, when choosing expressions from the Pareto front based on training NMSE, both ID and OOD values are consistently the lowest. Similarly, when choosing the expression in the Pareto front having the lowest complexity, the average complexity values across problems and runs are the lowest (see Figure \ref{fig:pareto_front_complexity}). A less obvious comparison is shown in Figure \ref{fig:pareto_front_similarity}, where the three criteria are compared based on \(SA_{AST}\) and \(SA_{Term}\). The similarity values remain low across all the three criteria, and no winner emerges. 

\clearpage
\section{Qualitative comparison against LLM-based competitors}
\label{app:llmsr_lasr_comparison}

Our approach shares with LLM-SR~\citep{shojaee2025llmsr} the general idea of using an LLM to explore the space of symbolic expressions within a loop while relying on data-driven evaluation to guide subsequent generations. Both methods also decouple structural search from coefficient optimization: the LLM proposes equation structures, while numerical coefficients are fitted using an external optimizer. Nevertheless, the two approaches operationalize the search process differently.

LLM-SR maintains a dynamic experience buffer organized according to a multi-island strategy. Previously evaluated equation programs are stored and sampled to construct in-context demonstrations for subsequent LLM queries. Sampling is biased toward high-performing clusters and, within a selected cluster, toward shorter programs. In addition, poorly performing islands are periodically reset. Therefore, LLM-SR incorporates several evolutionary-search principles, including fitness-based sampling, diversity preservation, and iterative refinement, although these mechanisms primarily regulate which previously discovered complete programs are provided back to the LLM as experience.

Our framework instead embeds the LLM within an explicit population-based evolutionary loop. At each generation, individuals are selected from the current population and used as parents for LLM-driven variation. The resulting offspring are evaluated and subsequently compete for survival according to the multi-objective selection strategy. 
A further distinction of EvoMO-SR is that evolutionary information is maintained not only at the level of complete candidate expressions. During evolution, structural differences between a parent and its offspring are analyzed to identify newly introduced symbolic substructures. Promising substructures are stored in a separate archive and can subsequently be included as guidance when generating mutations of other candidate expressions. Consequently, the population and the substructure archive play complementary roles: the former maintains complete candidate solutions and drives evolutionary selection, whereas the latter provides a lightweight memory of reusable symbolic components discovered during the search.
Furthermore, EvoMO-SR provides a set of evolutionary-based parameters to customize the search, such as parents and offspring counts, survival selection, elitism, parent selection strategies (e.g., random, tournament selection, roulette), niching mechanisms (e.g., sharing, clearing, novelty, map-elites).
Table~\ref{tab:llmsr_comparison} summarizes the main methodological differences. These distinctions should not be interpreted as implying that LLM-SR is not evolutionary; rather, the two methods use evolutionary principles at different levels of the search process. LLM-SR primarily exploits fitness-guided management and reuse of complete equation programs, while our approach combines LLM-based variation with explicit population-level selection and an additional mechanism for extracting and reusing symbolic building blocks.

\begin{table}[]
    \centering
    \small
        \caption{High-level comparison between LLM-SR and our framework. Both approaches employ iterative LLM-based equation generation guided by previous evaluations, but they differ in how evolutionary information is represented and used. $^\dagger$LLM-SR performs fitness-biased selection of programs from its experience buffer for in-context prompting; here, ``explicit parent selection'' refers specifically to selecting individuals as parents in a population-based evolutionary algorithm.}
    \begin{tabular}{p{0.42\linewidth}cc}
        
        \textbf{Component} & \textbf{LLM-SR} & \textbf{Our method} \\
        \midrule
        LLM-based structural generation
            & \checkmark & \checkmark \\
        External coefficient optimization
            & \checkmark & \checkmark \\
        Data-driven iterative feedback
            & \checkmark & \checkmark \\
        Multiple maintained candidates
            & \checkmark & \checkmark \\
        Island-based experience management
            & \checkmark & -- \\
        Fitness-biased sampling of previous solutions
            & \checkmark & \checkmark \\
        Explicit parent selection
            & --$^\dagger$ & \checkmark \\
        Explicit parent--offspring survival selection
            & -- & \checkmark \\
        Configurable elitist survival
            & -- & \checkmark \\
        Multi-objective survival selection
            & -- & \checkmark \\
        % Explicit accuracy--complexity Pareto optimization
        %     & -- & \checkmark \\
        % Parent--offspring structural change extraction
            % & -- & \checkmark \\
        Separate archive of reusable substructures
            & -- & \checkmark \\
        Substructure-guided LLM mutation
            & -- & \checkmark \\
        
    \end{tabular}

    \label{tab:llmsr_comparison}
\end{table}

Our method is also related to LaSR~\citep{srwithlearnedconceptlibrary}, which combines LLM guidance with an explicit evolutionary SR algorithm. In contrast to LLM-SR, LaSR directly builds upon the population-based search implemented by PySR, maintaining multiple populations and combining conventional symbolic mutation and crossover with LLM-guided counterparts. The LLM-based operators are applied probabilistically and are conditioned on a dynamically learned library of concepts, allowing semantic information extracted during the search to influence subsequent evolutionary operations. The central difference with our approach concerns the representation and acquisition of the knowledge used to guide evolution. LaSR constructs a library of natural-language concepts by prompting an LLM to abstract patterns from high- and low-performing expressions. These concepts are subsequently evolved by the LLM and sampled to condition hypothesis initialization, mutation, and crossover. Our framework instead operates directly at the symbolic level: structural changes introduced between a parent and its offspring are identified and candidate mathematical substructures are extracted from these changes. Promising substructures are stored in a dedicated archive and can later be explicitly provided to the LLM when generating new offspring. Therefore, while both methods exploit information accumulated during evolution to bias subsequent search, LaSR transfers this information through LLM-generated semantic abstractions, whereas our method transfers explicit symbolic building blocks discovered during parent--offspring evolutions. The two mechanisms also differ in how the guidance information is maintained. In LaSR, individual textual concepts are not assigned an explicit data-driven fitness; concept evolution is primarily intended to generate potentially useful ideas that increase exploration. In our framework, archived substructures remain associated with the predictive performance of the offspring in which they were discovered, and the archive has a bounded capacity in which candidate structures compete according to this performance. Finally, both approaches account for predictive accuracy and expression simplicity, although they employ these criteria differently. LaSR extracts a Pareto frontier over dataset loss and syntactic simplicity, which is subsequently used during concept abstraction and for selecting high-quality expressions. In the multi-objective variant of our framework, NMSE and symbolic complexity instead constitute the objectives of the evolutionary process itself, with non-dominated sorting and crowding-based environmental selection directly determining population survival. Table~\ref{tab:lasr_comparison} summarizes the main similarities and differences between the two approaches. 

\begin{table}[]
\centering 
\small 
\caption{High-level comparison between LaSR and our framework. Both approaches combine explicit evolutionary symbolic regression with LLM-based guidance, but differ substantially in how knowledge acquired during evolution is represented and reused. LaSR abstracts high- and low-performing expressions into natural-language concepts, whereas our method explicitly extracts and archives symbolic substructures introduced through parent--offspring transformations.}
%$^\dagger$LaSR maintains and extracts a Pareto frontier according to dataset loss and syntactic simplicity; here, explicit multi-objective environmental selection refers specifically to using multiple objectives directly for population survival through non-dominated sorting and diversity-preserving selection.}
\begin{tabular}{p{0.45\linewidth}cc} \textbf{Component} & \textbf{LaSR} & \textbf{Our method} \\ \midrule Explicit population-based evolutionary search & \checkmark & \checkmark \\ Data-driven iterative feedback & \checkmark & %\checkmark \\ Complexity-aware search & \checkmark &
\checkmark \\ LLM-guided evolutionary mutation & \checkmark & \checkmark \\ Standard symbolic GP mutation/crossover & \checkmark & -- \\ LLM-based mutation & \checkmark & \checkmark \\ LLM-based crossover & \checkmark & -- \\
Reuse of information discovered during evolution & \checkmark & \checkmark \\ Guidance representation & Textual concepts & Symbolic substructures \\ 
%Parent--offspring structural change extraction & -- &
%\checkmark \\ Direct archive of mathematical building blocks & -- & \checkmark \\
Data-driven quality associated with archived guidance & Indirect & \checkmark \\ Explicit multi-objective survival selection & -- & \checkmark \\ \end{tabular}  \label{tab:lasr_comparison} \end{table}

DrSR \citep{wang2025drsr} extends LLM-SR with data-aware insights and the summarization and reuse of ideas. However, in contrast to EvoMO-SR, DrSR acquires information through LLM-based data interpretation and reflection. Moreover, similarly to LLM-SR, this method retains promising equations in an experience buffer, but does not perform accuracy--complexity Pareto survival. 
LLM-Meta-SR \cite{zhang2026llm} combines LLM-driven evolution with multi-objective survival selection, but with different targets. While EvoMO-SR evolves symbolic equations, LLM-Meta-SR generates selection operators that are evaluated through inner SR runs across multiple meta-training datasets. The multi-objective mechanism is used to balance downstream predictive performance and operator code length. 

\subsection{Comparison over time, calls and tokens}
We measured computation time (i.e., per-run wall-clock time), the number of API requests, and the number of input/output tokens by executing EvoMO-SR, LLM-SR and LaSR over the \texttt{oscillator1}. Each method was evaluated in three runs with distinct random seeds. Table \ref{tab:comp_calls} reports the mean and standard deviation over the three runs. The lowest runtime was obtained by LaSR, while LLM-SR was approximately 5.5x slower than EvoMO-SR. 
Given that the main evolutionary loop of LaSR is managed by pySR and the GP operators are occasionally alternated with LLM-driven ones, the number of API requests is small compared to EvoMO-SR.
For LLM-SR, the number of API requests per run is equal to 250 because at each iteration the LLM is prompted to generate 4 distinct equations (thus requiring 1'000 model generations).
LLM-SR also generated considerably longer responses, producing
approximately 3.4 times as many output tokens as EvoMO-SR. For LaSR, the reduced overall costs are accompanied by lower predictive performance. Concerning LLM-SR, even though its numerical performance is similar to EvoMO-SR, its computation time is substantially longer.

\begin{table}[]
\centering
\caption{Mean per-run computation time (in seconds), number of API requests, number of input tokens and number of output tokens over the \texttt{oscillator1} dataset. Values are reported as mean $\pm$ standard
deviation over three runs.}
\label{tab:comp_calls}
\begin{tabular}{lcccc}

& \footnotesize \textbf{Runtime (s)} & \footnotesize \textbf{\# API requests} & \footnotesize \textbf{\# Input tokens}  & \footnotesize \textbf{\# Output tokens} \\ \hline
                \footnotesize LLM-SR            & \scriptsize$3'279.90 \pm 82.55$
& \scriptsize$250.00 \pm 0.00$   & \scriptsize$767'525.33 \pm 31'712.75$
& \scriptsize$506'200.33 \pm 1'565.21$ \\
\footnotesize LaSR              & \scriptsize$\mathbf{490.79 \pm 9.25}$
& \scriptsize$230.67 \pm 15.82$
& \scriptsize$59'708.67 \pm 3'734.57$
& \scriptsize$20'654.00 \pm 432.85$  \\     \hline
\footnotesize \textbf{EvoMO-SR} & \scriptsize$594.77 \pm 63.53$
& \scriptsize$1'000.00 \pm 0.00$
& \scriptsize $783'589.67 \pm 25'376.85$
& \scriptsize$150'259.00 \pm 7'743.26$ \\
\end{tabular}
\end{table}

\end{document}